\documentclass[10pt,twocolumn,letterpaper]{article}

\usepackage{multirow}
\usepackage[pagenumbers]{cvpr} 

\usepackage{xargs}
\usepackage[colorinlistoftodos,prependcaption,textsize=normalsize]{todonotes}
\newcommandx{\xiangyu}[2][1=]{\todo[linecolor=red,backgroundcolor=red!25,bordercolor=red,#1]{#2}}

\usepackage{times}
\usepackage{epsfig}
\usepackage{graphicx}
\usepackage{multirow}
\usepackage{caption}
\usepackage{multicol}
\usepackage{array}
\usepackage{booktabs}
\usepackage{threeparttable}
\definecolor{cvprblue}{rgb}{0.21,0.49,0.74}
\usepackage[pagebackref,breaklinks,colorlinks,citecolor=cvprblue]{hyperref}
\usepackage{color}
\usepackage{amsmath, amssymb}
\usepackage{algorithm}
\usepackage{algpseudocode}
\usepackage[capitalize]{cleveref}
\crefname{appendix}{Appx.}{Appxs.}
\crefname{section}{Sec.}{Secs.}
\Crefname{section}{Section}{Sections}
\crefname{table}{Table}{Tables}
\crefname{table}{Tab.}{Tabs.}

\newcolumntype{P}[1]{>{\centering\arraybackslash}p{#1}}

\definecolor{mygray}{gray}{.9}
\definecolor{ggray}{RGB}{127,127,127}
\definecolor{reda}{RGB}{192,0,0}
\definecolor{redb}{RGB}{217,148,143}
\definecolor{myyellow}{RGB}{190,144,0}
\definecolor{mygreen}{RGB}{80,100,40}
\definecolor{myblue}{RGB}{30,90,100}
\definecolor{dark-gray}{gray}{0.20}
\definecolor{middle-gray}{gray}{0.85}
\definecolor{light-gray}{gray}{0.93}

\definecolor{00blue}{RGB}{100,149,237}

\definecolor{lightblue}{rgb}{0.85, 0.95, 1}
\definecolor{lightorange}{rgb}{1, 0.95, 0.85}
\definecolor{lightpink}{rgb}{1, 0.9, 0.95}

\usepackage{epsfig}
\usepackage{graphicx}
\usepackage{amsmath}
\usepackage{amssymb}
\usepackage{multirow}
\usepackage{graphics}
\usepackage{threeparttable}
\usepackage{color}
\usepackage[normalem]{ulem}
\usepackage{float}
\usepackage{amsfonts}
\usepackage{bm}
\usepackage{bbm}
\usepackage{enumitem}
\usepackage{natbib}
\usepackage{subcaption}
\usepackage{array}
\usepackage{colortbl}
\usepackage{pifont}
\usepackage{booktabs}
\usepackage{mathtools}
\usepackage{siunitx}
\usepackage{caption} 
\usepackage{subcaption} 
\usepackage[nopar]{lipsum}
\usepackage{listings}
\usepackage[export]{adjustbox}
\usepackage{makecell}

\usepackage{mathtools}
\usepackage{siunitx}
\usepackage[nopar]{lipsum}
\usepackage{listings}

\usepackage{xargs}                      
\usepackage{xcolor}
\usepackage{wrapfig} 
\definecolor{myy}{RGB}{126,95,0}
\definecolor{mygray}{gray}{.9}
\definecolor{Gray}{gray}{0.9}
\definecolor{bblue}{RGB}{30,80,120}
\definecolor{mygray1}{gray}{.7}
\definecolor{ggray}{RGB}{127,127,127}
\definecolor{defaultcolor}{gray}{.9}
\definecolor{dark-gray}{gray}{0.20}

\definecolor{mygreen}{HTML}{39b54a}

\newcolumntype{x}[1]{>{\centering\arraybackslash}p{#1pt}}
\newcolumntype{y}[1]{>{\raggedright\arraybackslash}p{#1pt}}
\newcolumntype{z}[1]{>{\raggedleft\arraybackslash}p{#1pt}}

\newlength\savewidth

\usepackage{amsmath,amsfonts,bm}

\newcommand{\D}{\mD}
\newcommand{\F}{\mF}

\newcommand{\f}{\vf}
\newcommand{\x}{\vx}
\newcommand{\y}{\vy}
\newcommand{\z}{\vz}
\newcommand{\veps}{\bm{\varepsilon}}
\newcommand{\eps}{\bm{\epsilon}}
\newcommand{\dt}{\mathrm{d}t}
\newcommand{\dm}{\mathrm{d}}
\newcommand{\cskip}{c_{\text{skip}}}
\newcommand{\cin}{c_{\text{in}}}
\newcommand{\cout}{c_{\text{out}}}
\newcommand{\cnoise}{c_{\text{noise}}}
\newcommand{\cdata}{c_{\text{data}}}
\newcommand{\sigmad}{\sigma_{\text{d}}}

\def\eqref#1{equation~\ref{#1}}

\def\1{\bm{1}}

\def\eps{{\epsilon}}

\def\rvw{{\mathbf{w}}}

\def\vf{{\bm{f}}}

\def\vx{{\bm{x}}}
\def\vy{{\bm{y}}}
\def\vz{{\bm{z}}}

\def\mD{{\bm{D}}}

\def\mF{{\bm{F}}}

\DeclareMathAlphabet{\mathsfit}{\encodingdefault}{\sfdefault}{m}{sl}
\SetMathAlphabet{\mathsfit}{bold}{\encodingdefault}{\sfdefault}{bx}{n}

\def\gL{{\mathcal{L}}}

\def\gN{{\mathcal{N}}}

\def\gT{{\mathcal{T}}}
\def\gU{{\mathcal{U}}}

\newcommand{\E}{\mathbb{E}}

\definecolor{cvprblue}{rgb}{0.21,0.49,0.74}
\usepackage[pagebackref,breaklinks,colorlinks,citecolor=cvprblue]{hyperref}

\def\paperID{***} 
\def\confName{CVPR}
\def\confYear{2025}

\title{Transition Models: Rethinking the Generative Learning Objective}

\vspace{-5pt}
\author{
  \vspace{-25pt}\\
  {
  Zidong Wang$^{1,2,}$\thanks{: Equal contribution.
      $^\ddagger$: Project lead. 
      $^\dagger$: Corresponding authors: \href{mailto:bailei@pjlab.org.cn}{\color{black}{bailei@pjlab.org.cn}}, \href{mailto:liyangguang256@gmail.com}{\color{black}{liyangguang256@gmail.com}}.
  }, \quad
  Yiyuan Zhang$^{1,2,*, \ddagger}$, \quad
  Xiaoyu Yue$^{2,3}$, \quad
  Xiangyu Yue$^{1}$, \quad
  } \\ 
  {Yangguang Li$^{1,\dagger}$, \quad Wanli Ouyang$^{1,2}$, \quad Lei Bai$^{2,\dagger}$
  }
  \vspace{3pt} \\
  $^1$MMLab, CUHK ~~\quad\quad $^2$Shanghai AI Lab ~~\quad\quad $^3$USYD
  \vspace{3pt} \\
  \texttt{\small \{wangzd2022, yiyuanzhang.ai\}@gmail.com, \{xyyue, wlouyang\}@ie.cuhk.edu.hk} \\ 
  Code:~\, \url{https://github.com/WZDTHU/TiM}
  \vspace{-4pt} \\
}

\begin{document}

\maketitle

\begin{abstract}
A fundamental dilemma in generative modeling persists: iterative diffusion models achieve outstanding fidelity, but at a significant computational cost, while efficient few-step alternatives are constrained by a hard quality ceiling. This conflict between generation steps and output quality arises from restrictive training objectives that focus exclusively on either infinitesimal dynamics (PF-ODEs) or direct endpoint prediction. We address this challenge by introducing an exact, continuous-time dynamics equation that analytically defines state transitions across any finite time interval \(\Delta t\). This leads to a novel generative paradigm, Transition Models (TiM), which adapt to arbitrary-step transitions, seamlessly traversing the generative trajectory from single leaps to fine-grained refinement with more steps.
Despite having only 865M parameters, TiM achieves state-of-the-art performance, surpassing leading models such as SD3.5 (8B parameters) and FLUX.1 (12B parameters) across all evaluated step counts. Importantly, unlike previous few-step generators, TiM demonstrates monotonic quality improvement as the sampling budget increases. Additionally, when employing our native-resolution strategy, TiM delivers exceptional fidelity at resolutions up to \(4096\times4096\).
\end{abstract}

\section{Introduction}~\label{sec:intro}
Diffusion models have emerged as the dominant paradigm in visual content generation, producing state-of-the-art results across various domains~\cite{karras2022edm,peebles2023dit,podell2023sdxl,esser2024sd3,yang2024cogvideox,sora2024}. They generate samples from noise via iterative denoising, a process that can be formulated as numerical integration of either the reverse-time Stochastic Differential Equation (SDE) or the corresponding Probability-Flow Ordinary Differential Equation (PF-ODE), with related discrete-time solvers also widely used~\cite{song2020sde,song2020ddim,lu2022dpm}. Despite its effectiveness, iterative denoising often entails a large Number of Function Evaluations (NFEs)---approximately proportional to the number of integration steps—leading to increased inference latency and compute cost. 
\begin{figure}[t]
\begin{center}
\centerline{\includegraphics[width=0.97\linewidth]{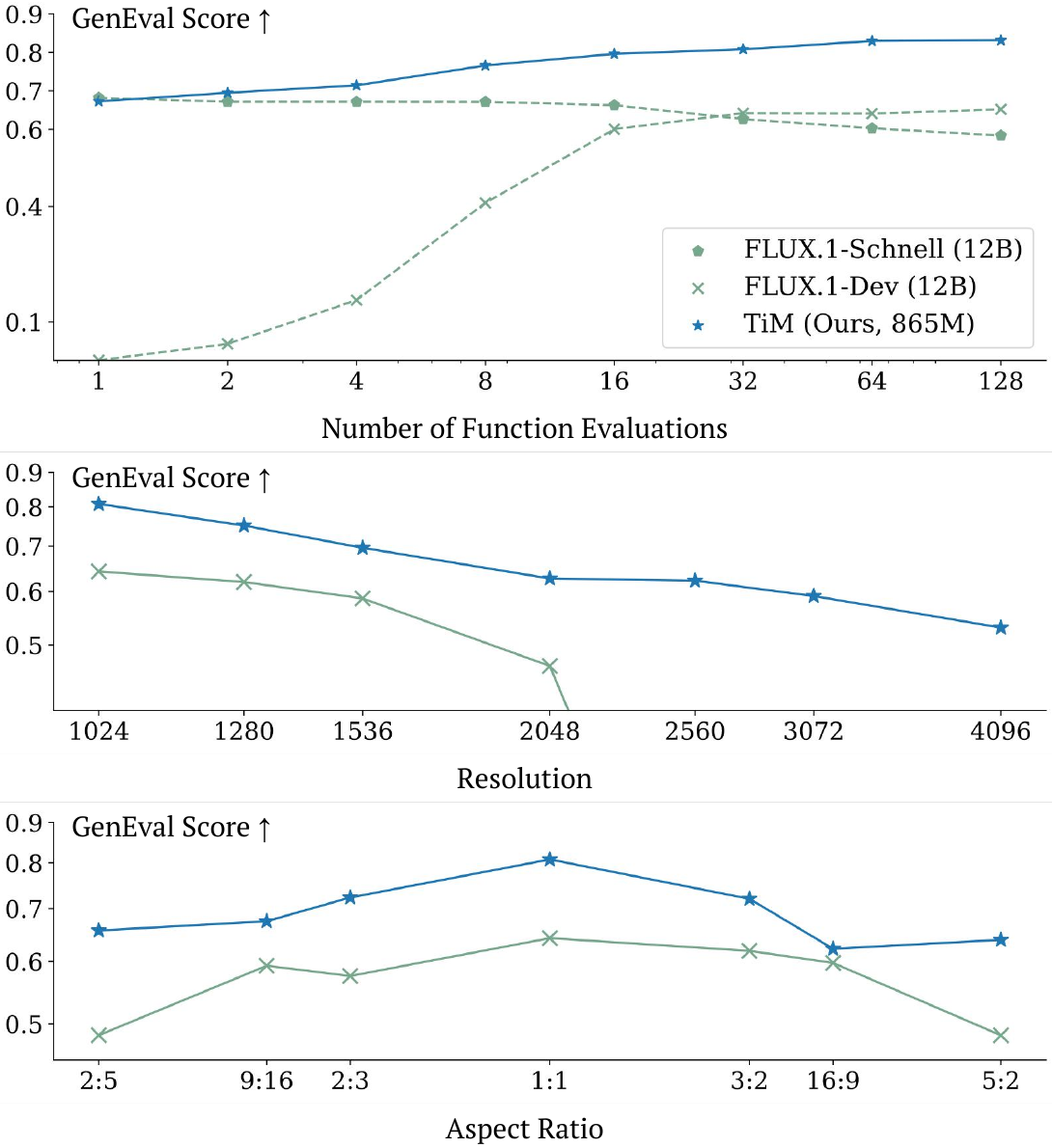}}
\vspace{-2mm}
\caption{\textbf{TiM's superior performance across different NFEs, resolutions, and aspect ratios.} On the GenEval~\cite{ghosh2023geneval} benchmark, TiM outperforms Flux.1 models~\cite{flux.1-dev2024,flux.1-sch2024} at different NFEs (top, \(1024\times1024\)), at higher resolutions (middle, \(1024\times1024\) to \(4096\times 4096\)), and diverse aspect ratios (bottom, $2:5$ to $5:2$).}
\vspace{-10mm}
\label{fig:comparison}
\end{center}
\end{figure}

In contrast, recent approaches reduce step counts by avoiding explicit multi-step integration. Consistency models~\cite{song2023cm,song2023icm,lu2024scm} impose PF-ODE self-consistency across different noise levels, while distribution-distillation methods~\cite{sauer2024sdxlturbo,lin2024sdxllightning,sauer2024sd3turbo,yin2024dmd,zhou2024sid} train students to approximate teacher distributions with fewer denoising steps. 
Shortcut~\cite{frans2024shortcut}, FlowMap~\cite{boffi2025flowmap,sabour2025ayflow}, and MeanFlow~\cite{geng2025meanflow,peng2025facm} learn the average (shortcut) velocity along the flow-matching trajectory via a self-consistency objective. The principle is that a single large step should approximate the integral of all smaller, instantaneous steps. However, by averaging the entire trajectory, \textit{they irrevocably discard the fine-grained local dynamics necessary for high-fidelity refinement}. This leads to performance saturation—while effective for few-step generation, it offers no gains from additional sampling budget. Moreover, despite these methods deliver strong few-step results; their compute–quality scaling is typically weaker than that of high-NFE diffusion models: \textit{quality gains plateau after only a few steps, and asymptotic performance remains below traditional multi-step diffusion}.

Thus, the entire field converges on a fundamental, yet flawed, compromise~\cite{ho2020ddpm,liu2023flow,song2023cm,yin2024dmd,geng2025meanflow}: models either achieve high fidelity at the cost of computational efficiency (\emph{e.g.}, diffusion models), or they gain efficiency by sacrificing the very dynamics needed for high-fidelity refinement (\emph{e.g.}, few-step models). The root of this dilemma is not architectural, but a learning objective. It stems from a foundational choice in how these models are taught to generate, not the specific components they are built from. This trade-off is a direct and inevitable consequence of the chosen \textit{granularity of supervision}. On one hand, local training methods that model instantaneous dynamics (such as those consistent with PF-ODEs/SDEs~\cite{ho2020ddpm,song2020sde,liu2023flow}) achieve high accuracy with small step sizes (\(\Delta t\)) and scale well to many-step generation. However, their performance degrades significantly in few-step regimes. On the other hand, finite-horizon training, which learns a direct mapping over a fixed interval (flow-map and consistency models~\cite{song2023cm,boffi2025flowmap,geng2025meanflow}), excels at few-step generation. Yet, these models see diminishing returns from additional intermediate steps unless specifically trained with complex, multi-interval objectives. This reveals a persistent dilemma: \textit{objectives that model instantaneous dynamics and those that learn finite-interval mappings each entail inherent limitations}. This motivates the question:
\textbf{What is an appropriate learning objective for generative models?}

We attempt to answer this
question from the following perspectives:

\textbf{1}) Diffusion training~\cite{lu2022dpm,lu2025dpm++,zheng2023dpm-v3} learns a local PF-ODE field whose numerical integration is accurate only in the small-step limit \(\Delta t\!\to\!0\). With large steps, the discretization error dominates; therefore, \textit{the objective should be flexible in terms of step sizes}.

\textbf{2}) Few-step objectives supervise an endpoint map but do not learn a compositional flow: without an approximate semigroup over time, extra steps change the path rather than refine it, causing schedule sensitivity and early saturation. Therefore, \textit{the objective requires a consistency along the trajectory, where intermediate steps act as refinements along a single trajectory, rather than deviations onto new ones}, which makes the sampler insensitive to step schedules and enables steady quality improvements with more steps.

Consequently, we argue that a generative model should learn a \textit{versatile denoising operator}, parameterized by the desired interval $\Delta t$. 
By learning the transitions between any state $\mathbf{x}_t$ to a previous state $\mathbf{x}_{t-\Delta t}$ for an \textit{arbitrary} $\Delta t$, the generative model is no longer approximating a differential equation or a statistic map. \textit{Instead, it is learning the solution manifold\footnote{solution manifold of a PF-ODE is the high-dimensional geometric surface formed by the collection of all possible generative trajectories that lead from noise to data.} of the generative process itself}. This formulation inherently unifies the local and finite-horizon perspectives, yielding a sampler that is both a powerful few-step generator and a precise, refinable integrator.
Since the training objective is to learn the transitions between any state to a previous state, thus, it is named as Transition Models (TiM), which parameterize state-to-state transitions along the PF-ODE trajectory for arbitrary time intervals.

We validate TiM's effectiveness through extensive experiments on text-to-image and class-guided image generation. As shown in \Cref{fig:comparison}, TiM shows superior performance across different NFEs, resolutions, and aspect ratios. On the GenEval~\cite{ghosh2023geneval} benchmark, our compact 865M parameter model, TiM, establishes a new state-of-the-art. It achieves a score of 0.67 with a single function evaluation (1-NFE) and scales to 0.83 at 128-NFE, outperforming billion-scale industrial models including SD3.5-Large~\cite{esser2024sd3} (8B) and FLUX.1-Dev~\cite{flux.1-dev2024} (12B).

\section{Related Work}

\noindent\textbf{Diffusion and Consistency Models.} 
Generative modeling has seen two dominant paradigms. 
Diffusion models~\cite{ho2020ddpm,karras2022edm,liu2023flow} iteratively solve a PF-ODE/SDE, achieving high quality but requiring many function evaluations (NFEs). 
In contrast, Consistency Models~\cite{song2023cm} learn a direct mapping for few-step generation but suffer from performance saturation and complex training requirements (e.g., pre-training and stabilization~\cite{song2023icm,lu2024scm}). 
While recent methods like FlowMap~\cite{boffi2025flowmap,sabour2025ayflow}  and MeanFlow~\citep{geng2025meanflow,peng2025facm} enable training CM-like models from scratch, they inherit the same limitation of stagnating quality with more steps. 

To break this impasse, we propose a new learning principle: mastering state transitions over arbitrary time intervals. This transforms the model from a brittle integrator or fixed-endpoint mapper into a robust navigator on the data manifold, preserving few-step efficiency while supporting monotonic refinement by using more steps.

\begin{figure*}[t]
    \centering
    \includegraphics[width=0.93\linewidth]{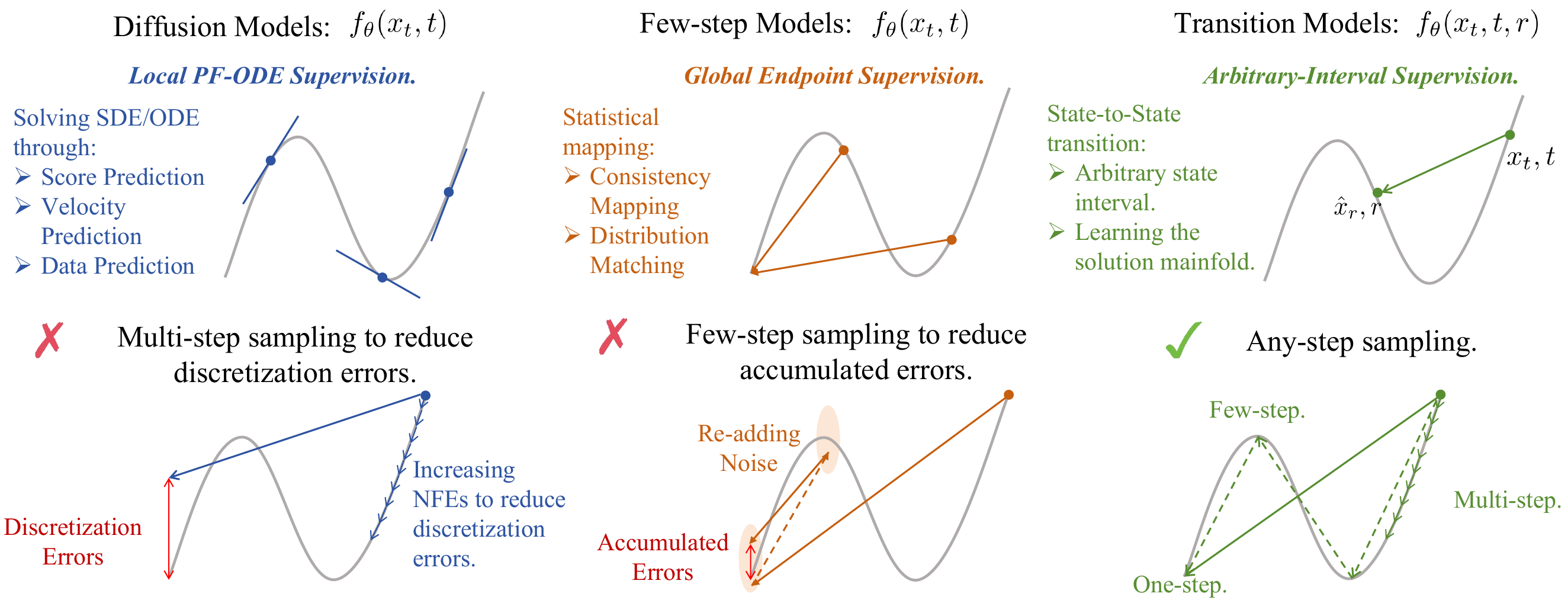}
    \caption{\textbf{Illustration of Different Generative Paradigms}. While conventional diffusion models learn the local vector field and few-step models learn a fixed endpoint map (a single large step), our Transition Models (TiM) are trained to master arbitrary state-to-state transitions. This approach allows TiM to learn the entire solution manifold of the generative process, unifying the few-step and many-step regimes within a single, powerful model.}
    \label{fig:illustration}
\end{figure*}

\noindent\textbf{Text-to-Image Generation with Few-steps.} 
Efficient text-to-image (T2I) sampling is currently dominated by distillation. These methods fall into two main camps: distribution distillation (e.g., SD-Turbo~\citep{sauer2024sdxlturbo,sauer2024sd3turbo}, DMD~\citep{yin2024dmd,yin2025dmd2}), which matches the teacher's output distribution, and trajectory distillation (e.g., LCM~\citep{luo2023lcm}, PCM~\cite{wang2025pcm}), which mimics its generation path. Hybrid methods~\citep{ren2025hypersd} combine both. 

However, all these approaches are fundamentally limited: \textbf{1}) they require a large, pre-trained teacher, leading to complex and costly pipelines, and \textbf{2}) they produce brittle, few-step-only models whose quality stagnates or degrades with more steps. We bypass these limitations entirely by introducing TiM, the first T2I generator trained from scratch that masters arbitrary-step sampling, delivering strong few-step results that monotonically improve with more compute.

\section{Transition Models}~\label{sec:method}
In this section, we first analyze the limitations of PF-ODE supervision in diffusion models, which constrain the state transition to a local, infinitesimal interval. 
To address the limitations, we generalize diffusion's local state transition to an arbitrary-interval state transition, as illustrated in \cref{fig:illustration}, from which we we derive a novel mathematical identity that links the state $\mathbf{x}_t$, the interval $\Delta t$, and the network $\boldsymbol{f}_\theta$. From this identity, we formulate a training objective governing state evolution over any interval $\Delta t$, and further propose two theoretically motivated improvements for scalable and stable training.
Finally, we present the architecture improvements for effective transition modeling. 

\subsection{Limitation of PF-ODE Supervision}~\label{sec:method:odes}
Given the noise distribution $\veps\sim\gN(\mathbf{0},\mathbf{I})$ and the data distribution $\x \sim p_\text{data}(\x)$, diffusion models learn to map the noise distribution to the data distribution. 
Given time range $t\in[0,T]$, the forward process utilizes coefficients $\alpha_t$ and $\sigma_t$, such that $\x_t=\alpha_t \x + \sigma_t \veps$, which can be described by an SDE~\cite{song2020sde}:
\begin{equation}
    \dm\x_t = \mathrm{f}(\x_t,t)\dt+\mathrm{g}(t)\dm\rvw,
\end{equation}
where $\rvw$ is the standard  Wiener process, $\mathrm{f}(\x_t,t) = \frac{\dot{\alpha}_t}{\alpha_t}\x_t$ is the drift coefficient and $\mathrm{g}(t)=2\sigma_t\dot{\sigma}_t-2\frac{\dot{\alpha}_t}{\alpha_t}\sigma_t^2$ is the diffusion coefficient~\cite{song2020sde,karras2022edm,lu2024scm,sun2025ucgm}.  \citet{anderson1982reverse-time-sde} and \citet{song2020sde} have shown that the forward process can be reversed by solving the reverse-time SDE from or equivalently the probability flow ODE (PF-ODE)\footnote{\citet{song2020sde} have shown that the PF-ODE trajectory has the same marginal probability as the reverse-time SDE: $\dm\x_t = [\mathrm{f}(\x_t,t)-\frac{1}{2}\mathrm{g}(t)^2\nabla_{\x_t}\log p_t(\x_t)]\dt + \mathrm{g}(t)\dm\rvw$.}:
\begin{equation}
\begin{aligned}
    \frac{\dm\x_t}{\dt}
    &=\mathrm{f}(\x_t,t)-\frac{1}{2}\mathrm{g}(t)^2\nabla_{\x_t}\log p_t(\x_t)\\
    &=\frac{\dm\alpha_t}{\dt}\x+\frac{\dm\sigma_t}{\dt}\veps,
\end{aligned}
\end{equation}
where $\nabla_{\x_t}\log p_t(\x_t)=-\frac{\veps}{\sigma_t}$ denotes the score function. 

Thus, a diffusion model can be parameterized as $\f_\theta(\x_t,t)=\F_\theta(\x_t,\cnoise(t))$, where $\theta$ denotes the parameters of the neural network and $\cnoise(t)$ is the time scaling function. 
The training objective can be given by:
\begin{equation}
    \E_{\x,\veps,t}[w(t)d(\f_\theta(\x_t,t)-(\hat{\alpha}_t\x+\hat{\sigma}_t\veps))],
    \label{eq:diffusion_objective}
\end{equation}
where $\hat{\alpha}_t$ and $\hat{\sigma}_t$ are the coefficients of diffusion target, $w(t)$ is a weighting function, $d(\cdot,\cdot)$ is a metric function such as the L2 loss $d(\x,\y)=\|\x-\y\|_2^2$. 

Despite different transports~\footnote{For convenience, we elaborate the coefficients $\alpha_t$, $\sigma_t$, $\hat{\alpha}_t$, and $\hat{\sigma}_t$ of different diffusion transports in \cref{tab:tim_formulation}} have instantiate coefficients $\alpha_t$ and $\sigma_t$, the training objectives are equivalent to supervising the PF-ODE field~\footnote{For example, in VE-SDE~\cite{song2020sde},with coefficients $\alpha_t=1,\sigma_t=t$, the PF-ODE is: $\frac{\dm\x_t}{\dt}=\veps$, and the training objective is $-\veps$. In OT-FM~\cite{lipman2022flowmatching}, with $\alpha_t=1-t,\sigma_t=t$, the PF-ODE is: $\frac{\dm\x_t}{\dt}=\veps-\x$, which directly matches the training objective.}. 
During sampling, diffusion models solve this PF-ODE, integrated from $t=T$ to $t=0$ using numerical solvers. To reduce discretization error and preserve the learned continuous-time dynamics, practical solvers~\cite{song2020ddim,lu2025dpm++,zheng2023dpm-v3} typically require a small step size (i.e., $\Delta t\rightarrow0$) or many sub-steps per interval (i.e., high-order solvers), thus inducing huge NFEs.

\subsection{State Transition}
The derivation begins with the general mathematical form for a state transition between points $(\boldsymbol{x}_t, \boldsymbol{x}_r)$ on a PF-ODE trajectory, as given in \cref{eq:tim_transition}. 
\textit{The central principle is to treat this form not as a numerical approximation, but as an exact identity that must hold for any interval $\Delta t = t-r$}. It allows us to formulate a \textit{general state transition dynamic} (\cref{eq:tim_identity}) that is valid across any interval. 
Consequently, the model's training objective is no longer constrained to approximating a local solution of the PF-ODE. 
Instead, it is trained to learn the \textit{entire solution manifold of the generative process}. 
By internalizing this global structure, the model inherently acquires the ability to perform inference over arbitrary step sizes, from large, single leaps to fine-grained, iterative refinement. We illustrate our derivation process step-by-step as follows:

\noindent\textbf{State Transition.} Given noisy state $\x_t=\alpha_t\x+\sigma_t\veps$, a diffusion model $\f_\theta(\x_t,t)$ is optimized towards the target $\hat{\alpha}_t\x+\hat{\sigma}_t\veps$, leading to the $x$-prediction and $\veps$-prediction: 
\begin{equation}
\hat{x}=\frac{\hat{\sigma}_t\x_t-\sigma_t\f_\theta(\x_t,t)}{\hat{\sigma}_t\alpha_t-\hat{\alpha}_t\sigma_t}, \quad
\hat{\veps}=\frac{\alpha_t\f_\theta(\x_t,t) - \hat{\alpha}_t\x_t}{\hat{\sigma}_t\alpha_t-\hat{\alpha}_t\sigma_t}.
\label{eq:x_and_eps_pred}
\end{equation}

Using the prediction $\hat{x}$ and $\hat{\veps}$, arbitrary previous state $\x_r$ ($r<t$) can be represented as:
\begin{equation}
\begin{aligned}
    \x_r 
    & = \alpha_r\hat{\x} + \sigma_r\hat{\veps} \\
    & = \frac{(\alpha_r\hat{\sigma}_t-\sigma_r\hat{\alpha}_t)\x_t + (\sigma_r\alpha_t-\alpha_r\sigma_t)\f_\theta(\x_t,t)}{\hat{\sigma}_t\alpha_t-\hat{\alpha}_t\sigma_t}.
\end{aligned}
\label{eq:state_transition}
\end{equation}
This represents the general form of a first-order state transition on the PF-ODE Trajectory.

\noindent\textbf{State Transition Identity.} Different from the diffusion model, our transition model learns the state transition $\f_\theta(\x_t,t,r)=\F_\theta(\x_t,\cnoise(t),\cnoise(r))$ between state $\x_t$ and $\x_r$. By introducing $\f_\theta(\x_t,t,r)$ to \cref{eq:state_transition}, we obtain:
\begin{equation}
    \x_r = \frac{(\alpha_r\hat{\sigma}_t-\sigma_r\hat{\alpha}_t)\x_t + (\sigma_r\alpha_t-\alpha_r\sigma_t)\f_\theta(\x_t,t,r)}{\hat{\sigma}_t\alpha_t-\hat{\alpha}_t\sigma_t}.
    \label{eq:tim_transition}
\end{equation}
Here, we define $A_{t,r}:=\frac{\alpha_r\hat{\sigma}_t-\sigma_r\hat{\alpha}_t}{\hat{\sigma}_t\alpha_t-\hat{\alpha}_t\sigma_t}$, $B_{t,r}:=\frac{\sigma_r\alpha_t-\alpha_r\sigma_t}{\hat{\sigma}_t\alpha_t-\hat{\alpha}_t\sigma_t}$, and $\f_{\theta,t,r}:=\f_\theta(\x_t,t,r)$ for simplicity, then we differentiate both sides with respect to $t$ and rearranging yields:
\begin{equation}
\begin{aligned}
    & \frac{\dm\x_r}{\dt}=\frac{\dm}{\dt}(A_{t,r}\x_t+B_{t,r}\f_{\theta,t,r}) 
    \Longrightarrow
    \\
    &\x_t\frac{\dm A_{t,r}}{\dt} + A_{t,r}\frac{\dm\x_t}{\dt} = -\f_{\theta,t,r}\frac{\dm B_{t,r}}{\dt} - B_{t,r}\frac{\dm\f_{\theta,t,r}}{\dt},
\end{aligned}
\label{eq:tim_differentiate}
\end{equation}
which can be further simplified as follows (detailed in \cref{appx:tim_derivation}):
{\small
\begin{equation}
\begin{aligned}
    & \frac{\dm(B_{t,r}\cdot(\hat{\alpha}_t\x + \hat{\sigma}_t\veps - \f_{\theta,t,r}))}{\dt} = 0 \Longrightarrow 
    \\
    & (\underbrace{\hat{\alpha}_t\x + \hat{\sigma}_t\veps - \f_{\theta,t,r}}_{\text{PF-ODE supervision}})\frac{\dm B_{t,r}}{\dt} +  B_{t,r}\underbrace{\frac{\dm(\hat{\alpha}_t\x+\hat{\sigma}_t\veps -\f_{\theta,t,r})}{\dt}}_{\text{time‑slope matching}} = 0 .
\end{aligned}
\label{eq:tim_identity}
\end{equation}
}

We denote \Cref{eq:tim_identity} as the State Transition Identity, a product-derivative invariant. 
The State Transition Identity, $\frac{\mathrm{d}}{\mathrm{d}t}(B_{t,r} \cdot h(t)) = 0$, where $h(t) = \hat{\alpha}_t\boldsymbol{x} + \hat{\sigma}_t\boldsymbol{\varepsilon} - \boldsymbol{f}_{\theta,t,r}$ is the instantaneous residual, imposes a powerful two-fold constraint on the generative model $\boldsymbol{f}_\theta$.

\begin{itemize}
    \item \textbf{Implicit Trajectory Consistency:} The identity dictates that the weighted residual $B_{t,r}h(t)$ must be constant for any starting time $t$ leading to the same target $\boldsymbol{x}_r$. This directly enforces path consistency: the direct map $(t \to r)$ must be equivalent to any composition of intermediate steps, such as $(t \to s) \circ (s \to r)$. This property (\cref{eq:tim_identity} ), absent in standard consistency models, is the core mechanism that makes TiM robust to sampling schedules and enables monotonic refinement.

    \item \textbf{Time-Slope Matching:} Unpacking the product rule reveals that $(\frac{\mathrm{d}}{\mathrm{d}t}B_{t,r})h(t) + B_{t,r}(\frac{\mathrm{d}}{\mathrm{d}t}h(t)) = 0$. Unlike conventional diffusion training, which only minimizes the residual's value ($h(t) \to 0$), our objective forces the model to also minimize the residual's temporal derivative ($\frac{\mathrm{d}}{\mathrm{d}t}h(t) \to 0$). This higher-order supervision compels the model to learn a smoother solution manifold, preserving coherence during large-step sampling and ensuring stable refinement with smaller steps.
\end{itemize}

Derived from State Transition Identity (\cref{eq:tim_identity}), we obtain the learning target $\hat{\boldsymbol{f}}$ :
{\small
\begin{equation}
    \hat{\f} = \hat{\alpha}_t\x+\hat{\sigma}_t\veps+\frac{B_{t,r}}{\frac{\dm B_{t,r}}{\dt}}\left(\frac{\dm\hat{\alpha}_t}{\dt}\x+\frac{\dm\hat{\sigma}_t}{\dt}\veps-\frac{\dm\f_{\theta^-,t,r}}{\dt}\right),
    \label{eq:tim_target}
\end{equation}
}
where $\theta^-$ indicates the fixed network parameter $\theta$ and $\frac{\dm\f_{\theta^-,t,r}}{\dt}$ is the time derivative of the network.  

\subsection{Scalability and Stability in TiM Training}
\noindent\textbf{Remark 1: Making TiM Training Scalable.} 

A critical challenge in implementing our training target (\cref{eq:tim_target}) is the computation of the network's time derivative, $\frac{\mathrm{d}\boldsymbol{f}_{\theta^-,t,r}}{\mathrm{d}t}$. 
Prior work, such as MeanFlow~\citep{geng2025meanflow,sabour2025ayflow,peng2025facm} and sCM~\cite{lu2024scm}, relies on the Jacobian-Vector Product (JVP) for this task. 
However, JVP presents a \textit{fundamental roadblock to scalability}. 
It is not only compute-intensive but, more cripplingly, its reliance on backward-mode automatic differentiation is \textbf{incompatible with essential training optimizations}, including FlashAttention~\citep{dao2023flashattention-2} and distributed frameworks of FSDP~\citep{zhao2023fsdp}. 
This incompatibility has effectively rendered JVP-based methods impractical for training billion-parameter foundation models.

We break this barrier with the \textbf{Differential Derivation Equation (DDE)}, a principled and highly efficient finite-difference approximation:
\begin{equation}
    \frac{\mathrm{d}\boldsymbol{f}_{\theta^-,t,r}}{\mathrm{d}t} \approx \frac{\boldsymbol{f}_{\theta^-}(\boldsymbol{x}_{t+\epsilon},t+\epsilon,r)-\boldsymbol{f}_{\theta^-}(\boldsymbol{x}_{t-\epsilon},t-\epsilon,r)}{2\epsilon}.
    \label{eq:dde_calculation}
\end{equation}
As shown in \cref{tab:derivative}, DDE is not only $\sim$2$\times$ faster than JVP but, crucially, its forward-pass-only structure is \textit{natively compatible with FSDP}. 
This compatibility transforms a previously unscalable training process into one ready for large-scale deployment, making TiM the first model of its kind practical for from-scratch, billion-parameter pre-training~\footnote{We provide a detailed analysis of DDE in the Appendix \cref{tab:dde_epsilon}}.

\begin{table}[t]
\centering
\resizebox{\linewidth}{!}{
\begin{tabular}{c|cc|cc|ccc}
\toprule
\multirow{2}*{Method} & 
\multicolumn{2}{c|}{Operator} & 
\multicolumn{2}{c|}{Training} & 
\multicolumn{3}{c}{FID} 
\\
& FLOPs & Latency  & Throughput  & Memory  & NFE=1 & NFE=8 & NFE=50 \\
& (G) & (ms) & (/s) & (GiB) \\
\midrule
JVP & 48.29 & 213.14 & 1.80 & 14.89 & \textbf{49.75} & 26.22 & 18.11 \\
DDE & \textbf{24.14} &\textbf{ 110.08} & \textbf{2.40} & 15.23 & 49.91 & \textbf{26.09} & \textbf{17.99} \\
\bottomrule
\end{tabular}
}
\caption{\textbf{Derivative Calculation Comparison.} We utilize a TiM-B/4 model for latency, throughput, and memory measurement, with a batch size of 256 on a NVIDIA-A100 GPU using BF16 precision.}
\label{tab:derivative}
\vspace{-5mm}
\end{table}

\begin{table*}[t]
\centering
\begin{adjustbox}{max width=0.98\linewidth}
\begin{tabular}{lcc|ccccccc}
\toprule
Model & Param. & NFE & {Overall$\uparrow$} & {Single Obj.} & {Two Obj.} & {Counting} & {Colors} & {Position} & {Attr. Binding} \\
\midrule
\multicolumn{10}{l}{\textbf{\textit{Autoregressive Models}}} \\
Emu3-Gen~\cite{wang2024emu3}        & - & - & 0.54 & 0.98 & 0.71 & 0.34 & 0.81 & 0.17 & 0.21 \\
GPT-4o~\cite{achiam2023gpt4}        & - & - & {0.84} & 0.99 & 0.92 & 0.85 & 0.92 & 0.75 & 0.61 \\
\midrule
\multicolumn{8}{l}{\textbf{\textit{Multi-step Diffusion Models}}} \\
SD2.1~\cite{rombach2022ldm}           
& 865M & 100 & 0.50 & 0.98 & 0.51 & 0.44 & 0.85 & 0.07 & 0.17 \\
SD-XL~\cite{podell2023sdxl}   
& 2.6B & 100 & 0.55 & 0.98 & 0.74 & 0.39 & 0.85 & 0.15 & 0.23 \\
Seedream2.0~\cite{gong2025seedream2}
& - & - & 0.84 & 1.0 & 0.98 & 0.91 & 0.94 & 0.47 & 0.75 \\
SD3.5-Medium~\cite{esser2024sd3}
& 2B & 100 & 0.63 & 0.98 & 0.78 & 0.50 & 0.81 & 0.24 & 0.52 \\
SD3.5-Large~\cite{esser2024sd3}
& 8B & 128 & 0.69 & 0.99 & 0.89 & 0.67 & 0.81 & 0.24 & 0.56 \\
SANA-1.5~\cite{xie2025sana1.5}
& 4.6B & 40 & 0.81 & 0.99 & 0.93 & 0.86 & 0.84 & 0.59 & 0.65 \\
FLUX.1-Dev~\cite{flux.1-dev2024}
& 12B & 128 & 0.65 & 0.98 & 0.79 & 0.69 & 0.76 & 0.21 & 0.48 \\
\midrule
\multicolumn{10}{l}{\textbf{\textit{Few-step Distilled Diffusion Models}}} \\
SDXL-LCM~\cite{luo2023lcm}
& 2.6B & 8 & 0.40 & 0.97 & 0.50 & 0.12 & 0.67 & 0.09 & 0.07\\
SDXL-Turbo~\cite{sauer2024sdxlturbo}
& 2.6B & 8 & 0.50 & 0.99 & 0.75 & 0.07 & 0.89 & 0.11 & 0.20 \\
Hyper-SDXL~\cite{ren2025hypersd}
& 2.6B & 8 & 0.46 & 0.92 & 0.58 & 0.26 & 0.78 & 0.11 & 0.15 \\
SANA-Sprint~\cite{chen2025sana-sprint}
& 1.6B & 8 & 0.72 & 1.0 & 0.88 & 0.56 & 0.87 & 0.56 & 0.47 \\
SD3.5-Turbo~\cite{sauer2024sd3turbo}
& 8B & 8 & 0.66 & 0.99 & 0.81 & 0.62 & 0.79 & 0.25 & 0.48 \\
FLUX.1-Schnell~\cite{flux.1-dev2024}
& 12B & 1 & 0.68 & 0.99 & 0.88 & 0.63 & 0.78 & 0.27 & 0.53 \\
\midrule
\multicolumn{10}{l}{\textbf{\textit{Transition Models}}} \\
\rowcolor{gray!20} 
&  & 1 & 0.67 & 0.98 & 0.75 & 0.52 & 0.80 & 0.54 & 0.44 \\
\rowcolor{gray!20}
TiM & 865M                     & 8 & 0.76 & 0.99 & 0.87 & 0.61 & 0.88 & 0.63 & 0.61 \\
\rowcolor{gray!20}
                    &      & 128 & 0.83 & 1.0 & 0.91 & 0.73 & 0.91 & 0.73 & 0.71 \\
\bottomrule
\end{tabular}
\end{adjustbox}
\caption{\textbf{System-level quality comparison of TiM and SOTA methods on GenEval benchmark.} In the table, 1-NFE denotes a single sampling step; 8-NFE corresponds to four sampling steps with CFG, and other multi-NFE follow the same convention. Compared with multi-step diffusion models and few-step distilled models, TiM offers any-step generation, delivering strong few-step performance and exhibiting consistent, stable improvements as NFE increases.}
\label{tab:geneval_main}
\vspace{-3mm}
\end{table*}

\noindent\textbf{Remark 2: Making TiM Training Stable.} 

In addition to scalability, a key challenge in training with arbitrary intervals is managing gradient variance. For example, transitions over very large intervals $(\Delta t \to t)$ are easier to make loss spikes. 
To mitigate this, we introduce a loss weighting scheme that prioritizes short-interval transitions, which are more frequent and provide a more stable learning signal.

The weighting function, $w(t,r)$, is a composition of a time-warping function $\tau(\cdot)$ and a kernel function $k(\cdot, \cdot)$:
\begin{equation}
    w(t,r) = k(\tau(t), \tau(r)).
    \label{eq:weight_fn}
\end{equation}
Here, $\tau(\cdot)$ is a monotonic function that re-parameterizes the time axis.
For our final model, we use a tangent space transformation, which effectively stretches the time domain, yielding the specific weighting:
\begin{equation}
    w(t,r) = ({\sigma_{\text{data}} + \tan(t) - \tan(r)})^{-\frac{1}{2}} ,
\end{equation}
where $\sigma_{\text{data}}$ is the standard deviation of the clean data~\footnote{In the Appendix, we conduct an in-depth comparison of alternative weighting schemes is provided in \cref{tab:time_weighting}.}.

\noindent\textbf{Learning Objective.}
Our theoretical framework culminates in a scalable and stable learning objective. 
We train the network $\boldsymbol{f}_\theta$ to predict the dynamic target $\hat{\boldsymbol{f}}$ in \cref{eq:tim_target}. 
To manage gradient variance and ensure stable convergence, this is weighted by the interval function $w(t,r)$ from \cref{eq:weight_fn}. 
This results in the final TiM objective:
\begin{equation}
\mathbb{E}_{\boldsymbol{x},\boldsymbol{\varepsilon},t,r}\left[w(t,r) \cdot d\left(\boldsymbol{f}_\theta(\boldsymbol{x}_t, t, r) - \hat{\boldsymbol{f}}\right)\right].
\label{eq:tim_objective}
\end{equation}
This objective generalizes the standard PF-ODE supervision to arbitrary state-to-state transitions. 
The practical implementation, enabled by our efficient DDE calculation, is detailed in \cref{alg:tim_train}. 
We summarize the specific parameterizations for various transport choices in \cref{tab:tim_formulation}.

\subsection{Improved Architectures}
\label{subsec:method_architecture}
We conduct a series of experiments to explore architectural modifications based on DiT~\cite{peebles2023dit} for effective state transition learning in \cref{tab:architecture}. We illustrate the exact architectures in the Appendix \ref{appx:tim_framework}.

\noindent\textbf{Decoupled Time and Interval Embeddings.}
To enable the model to distinguish between the absolute time $t$ and the transition interval $\Delta t$, we introduce a decoupled embedding strategy. 
We employ two independent time encoders, $\phi_t$ and $\phi_{\Delta t}$, to parameterize these two quantities. 
Their outputs are summed to form the final time-conditioning vector:
\begin{equation}
    \mathbf{E}_{t, \Delta t} = \phi_t(t) + \phi_{\Delta t}(\Delta t).
    \label{eq:time_embed}
\end{equation}
This time embedding is then integrated with task-specific conditioning as follows:
\begin{itemize}
    \item {For class-guided generation,} the class embedding $\mathbf{E}_c$ is added to the time embedding, and the resulting sum, $\mathbf{E}_{t, \Delta t} + \mathbf{E}_c$, modulates the AdaLN layers of the model.
    \item {For text-to-image generation,} the conditioning pathways are separated. The time embedding $\mathbf{E}_{t, \Delta t}$ solely modulates the AdaLN layers, while textual features from the prompt are injected via dedicated cross-attention mechanisms.
\end{itemize}

\noindent\textbf{Interval-Aware Attention.}
We assume that the optimal way to model spatial dependencies is conditional on the transition interval $\Delta t$. 
A large step $(\Delta t \to t)$ may require global, coarse-grained restructuring, while a small step $(\Delta t \to 0)$ demands fine-grained, local refinement. 
Standard self-attention, which is agnostic to this context, is inappropriate for this task.
We therefore introduce the {Interval-Aware Attention}, a mechanism that infuses the transition interval's magnitude directly into the query, key, and value computations. 
Specifically, we project both the spatial tokens $\mathbf{z}$ and the interval embedding $\mathbf{E}_{\Delta t}$ into a shared representational space before the attention calculation:
\begin{equation}
\begin{aligned}
    \mathbf{q} &= \mathbf{z}\mathbf{W}_q + \mathbf{b}_q + \mathbf{E}_{\Delta t}\mathbf{W}'_q, \\
    \mathbf{k} &= \mathbf{z}\mathbf{W}_k + \mathbf{b}_k + \mathbf{E}_{\Delta t}\mathbf{W}'_k, \\
    \mathbf{v} &= \mathbf{z}\mathbf{W}_v + \mathbf{b}_v + \mathbf{E}_{\Delta t}\mathbf{W}'_v.
\end{aligned}
\label{eq:distance_aware_attn}
\end{equation}
Here, $(\mathbf{W}_q, \mathbf{W}_k, \mathbf{W}_v)$ are the primary projection matrices for the spatial tokens, while $(\mathbf{W}'_q, \mathbf{W}'_k, \mathbf{W}'_v)$ are dedicated projection matrices that modulate the attention based on the interval embedding.
\section{Experiments}

\begin{table*}[t]
\centering
\begin{adjustbox}{max width=.95\textwidth}
\begin{tabular}{lcc|cc|cccccc}
\toprule
\multirow{2}*{Model} & \multirow{2}*{Param.} & \multirow{2}*{NFE} & \multicolumn{2}{c|}{MJHQ30K} & \multicolumn{6}{c}{DPGBench} \\
& & & {FID$\downarrow$} & {CLIP$\uparrow$} & {Overall$\uparrow$} & {Global} & {Entity} & {Attribute} & {Relation} & {Other} \\
\midrule
PixArt-$\alpha$~\cite{chen2023pixart-alpha} 
& 610M & 100 & 6.14 & 27.55 & 71.11 & 74.97 & 79.32 & 78.60 & 82.57 & 76.96 \\
PixArt-$\Sigma$~\cite{chen2024pixart-sigma} 
& 610M & 100 & 6.15 & 28.26 & 80.54 & 86.89 & 82.89 & 88.94 & 86.59 & 87.68 \\
SDXL~\cite{podell2023sdxl}            
& 2.6B & 100 & 6.63 & 29.03 & 74.65 & 83.27 & 82.43 & 80.91 & 86.76 & 80.41 \\
Playground v2.5~\cite{li2024playgroundv2.5}
& 2.6B & 100 & 6.09 & 29.13 & 75.47 & 83.06 & 82.59 & 81.20 & 84.08 & 83.50 \\
Hunyuan-DiT~\cite{li2024hunyuan-dit}
& 1.5B & 100 & 6.54 & 28.19 & 78.87 & 84.59 & 80.59 & 88.01 & 74.36 & 86.41 \\
SD3.5-Medium~\cite{esser2024sd3}    & 2B & 100 & 11.92 & 27.83 & 84.08 & 87.90 & 91.01 & 88.83 & 80.70 & 88.68 \\
SD3.5-Turbo~\cite{sauer2024sd3turbo}     
& 8B & 8 & 11.97 & 27.35 & 79.03 & 80.12 & 86.13 & 84.73 & 91.86 & 78.29 \\
SD3.5-Large~\cite{esser2024sd3}     
& 8B & 32 & 14.68 & 27.88 & 83.21 & 84.27 & 88.99 & 87.35 & 93.28 & 80.35 \\
FLUX.1-Schnell~\cite{flux.1-sch2024}  
& 12B & 8 & 7.94 & 28.14 & 84.94 & 86.62 & 90.82 & 88.35 & 93.45 & 82.00 \\
FLUX.1-dev~\cite{flux.1-dev2024}      
& 12B & 32 & 9.19 & 27.27 & 83.32 & 81.46 & 90.02 & 87.50 & 92.72 & 82.39 \\ \midrule
\rowcolor{gray!20} 
 & & 1 & 6.68 & 24.80 & 74.93 & 82.98 & 83.64 & 83.54 & 91.99 & 63.20 \\
\rowcolor{gray!20} 
\textbf{TiM}            & 865M  & 8 & 5.28 & 26.10 & 81.30 & 82.01 & 88.31 & 87.81 & 93.37 & 70.80 \\
\rowcolor{gray!20} 
                        &      & 32 & 5.65 & 26.31 & 82.71 & 82.67 & 89.40 & 88.48 & 93.31 & 79.20 \\
\bottomrule
\end{tabular}
\end{adjustbox}
\caption{\textbf{System-level quality comparison on MJHQ30K and DPGBench benchmarks.}}
\label{tab:mjhq_dpgbench}
\vspace{-3mm}
\end{table*}
\begin{table}[t]
\centering
\resizebox{\linewidth}{!}{
\begin{tabular}{lccc}
\toprule
Method & NFE=1 & NFE=8 & NFE=50 \\
\midrule
\multicolumn{4}{l}{\textit{Training Objective}} \\
(a) Baseline (SiT-B/4~\cite{ma2024sit}) & 309.5 & 77.26 & {20.35} \\
(b) TiM-B/4 (\textit{w/} JVP) & \textbf{49.75} & 26.22 & 18.11 \\
(c) TiM-B/4 (\textit{w/} DDE) & 49.91 & \textbf{26.09} & \textbf{17.99} \\
\midrule
\multicolumn{4}{l}{\textit{Architecture}} \\
(d) Vanilla Architecture                    & 56.22 & 28.75 & 20.37 \\
(e) + Decoupled Time Embedding (De-TE)      & 49.91 & 26.09 & 17.99 \\
(f) + Interval-Aware Attention (IA-Attn)    & 48.38 & 26.10 & 17.85 \\
(g) + De-TE + IA-Attn                       & \textbf{48.30} & \textbf{25.05} & \textbf{17.43} \\
\midrule
\multicolumn{4}{l}{\textit{Training Strategy} (on top of (g))} \\
(h) + Time-weighting & \textbf{47.46} & \textbf{24.62} & \textbf{17.10} \\
\bottomrule
\end{tabular}
}
\caption{\textbf{Ablation studies of Transition Models on the standard ImageNet-256 benchmark (FID\(\downarrow\))}. We analyze the effect of training objectives, architecture, and training strategies.}
\label{tab:architecture}
\end{table}

\begin{table}[t]
\centering
\resizebox{\linewidth}{!}{
\begin{tabular}{lcccc}
\toprule
{Method} & NFE=1 & NFE=8 & NFE=32 & NFE=128 \\
\midrule
SD3.5-Turbo~\cite{sauer2024sd3turbo}     
& 0.50 & 0.66 & 0.70 & 0.70 \\
FLUX.1-Schnell~\cite{flux.1-sch2024}  
& 0.68 & 0.67 & 0.63 & 0.58 \\
SD3.5-Large~\cite{esser2024sd3}
& 0.00 & 0.50 & 0.69 & 0.70 \\
FLUX.1-Dev~\cite{flux.1-dev2024}
& 0.00 & 0.40 & 0.64 & 0.65 \\ \midrule
\rowcolor{gray!20} 
\textbf{TiM} & 0.67 & 0.76 & 0.80 & 0.83\\
\bottomrule
\end{tabular}
}
\caption{\textbf{Benchmarking generation quality across NFEs on the GenEval benchmark (score\(\uparrow\)).} We compare a single TiM model against diffusion models (i.e., SD3.5-Large and FLUX.1-Dev) and distilled models (i.e., SD3.5-Turbo and FLUX.1-Schnell).}
\label{tab:nfe_comparison}
\vspace{-3mm}
\end{table}

\begin{table*}[t]
\centering
\begin{adjustbox}{max width=1.\textwidth}
\begin{tabular}{lc|cccccc|cccccc}
\toprule
\multirow{2}*{Method} & \multirow{2}*{NFE} & \multicolumn{6}{c|}{Aspect Ratio} & \multicolumn{6}{c}{Resolution} \\
& & $2:5$ & $9:16$ & $2:3$ & $3:2$ & $16:9$ & $5:2$ & $1280$ & $1536$ & $2048$ & $2560$ & $3072$ & $4096$ \\
\midrule
SD3.5-Turbo~\cite{sauer2024sd3turbo}
& 8  & \ding{55} & 0.53 & 0.60 & 0.58 & 0.30 & \ding{55} & 0.61 & \ding{55} & \ding{55} & \ding{55} & \ding{55} & \ding{55} \\
FLUX.1-Schnell~\cite{flux.1-sch2024}
& 8  & 0.57 & 0.61 & 0.63 & 0.62 & 0.59 & 0.57 & 0.64 & 0.58 & 0.46 & 0.14 & \ding{55} & \ding{55} \\ 
\rowcolor{gray!20} 
\textbf{TiM}      & 8  & 0.55 & 0.58 & 0.63 & 0.64 & 0.58 & 0.56 & 0.70 & 0.61 & 0.49 & 0.48 & 0.45 & 0.39 \\
\midrule
SD3.5-Large~\cite{esser2024sd3}
& 32 & 0.25 & 0.48 & 0.60 & 0.57 & 0.16 & \ding{55} & 0.63 & \ding{55} & \ding{55} & \ding{55} & \ding{55} & \ding{55} \\
FLUX.1-Dev~\cite{flux.1-dev2024}
& 32 & 0.48 & 0.59 & 0.62 & 0.60 & 0.59 & 0.57 & 0.62 & 0.58 & 0.49 & 0.27 & \ding{55} & \ding{55} \\
\rowcolor{gray!20} 
\textbf{TiM}      & 32 & 0.66 & 0.67 & 0.72 & 0.72 & 0.62 & 0.64 & 0.75 & 0.69 & 0.63 & 0.62 & 0.59 & 0.53 \\
\bottomrule
\end{tabular}
\end{adjustbox}
\caption{\textbf{Benchmarking resolution generation capabilities on GenEval Benchmark.} For aspect ratio generalization, the exact resolutions are: $\{1024\times2560, 1024\times1856, 1024\times1536, 1536\times1024, 1856\times1024, 2560\times1024\}$. \ding{55}: when GenEval score falls below $0.10$, we interpret it as evidence that the model fails to generalize to that resolution.}
\label{tab:resolution_comparison}
\vspace{-3mm}
\end{table*}

\subsection{Setup}
We use SD-VAE~\citep{rombach2022ldm} for ImageNet-$256\times256$ experiments and DC-AE~\citep{chen2024dc-ae} for text-to-image (T2I) experiments. Model architecture follows DiT~\citep{peebles2023dit}, except the modifications in \cref{subsec:method_architecture}. For T2I generation, we use $33M$ images from public datasets~\citep{sharma2018cc3m,changpinyo2021cc12m,megalith,schuhmann2021laion,t2i2m,chen2025blip3o,desai2021redcaps,singla2024pixelprose}. We train the T2I model with $865M$ parameters using the native-resolution training strategies for about 30 days using 16 NVIDIA-A100 GPUs. Gemma3-1B-it~\cite{team2025gemma3} is utilized as a text encoder. See more details in \cref{appx:training_details}.
We report the Number of Function Evaluations (NFE) to quantify sampling steps. When classifier-free guidance (CFG) is used, NFE doubles, because each step requires two model evaluations: one conditioned and one unconditioned.
We provide the T2I experiments and ablation experiments on ImageNet-$256\times256$ below and more results on class-guided image generation in \cref{appx:additional_ablations,appx:additional_c2i}.

\noindent\textbf{Native-Resolution Training.} 
Previous methods~\cite{wang2025nit,gong2025seedream2,gao2025seedream3} have shown the success of native-resolution training on resolution generalization; thus, we adopt this strategy for text-to-image generation, which preserves the original image resolution and aspect ratio information to the greatest extent possible. Given the wide resolution range, we increase noise for higher-resolution images and decrease it for lower-resolution ones. Following \citet{esser2024sd3}, we therefore apply resolution-dependent timestep shifting. Please see more details in \cref{appx:training_details}. 

\noindent\textbf{Sampling.} Since TiM learns the arbitrary state transition on the diffusion trajectory, it supports arbitrary-step sampling when producing images. Given a set of timesteps $\gT=\{t_i\}_{i=N}^{0}$ where $t_N=T, t_0=0$, we obtain the next state $\x_{t_{n-1}}$ given the current state $\x_{t_n}$ based on \cref{eq:state_transition}, as illustrated in ~\cref{alg:tim_sample}.

\subsection{Text-to-Image Generation}
TiM establishes a new state-of-the-art in performance, efficiency, and flexibility across diverse benchmarks (\cref{tab:geneval_main,tab:mjhq_dpgbench,tab:nfe_comparison,tab:resolution_comparison}). 
It achieves an SOTA FID of 5.25 on MJHQ-30K while resolving the core speed-quality trade-off. 
On GenEval, TiM's 1-NFE performance surpasses 8-NFE distilled models (e.g., SDXL-Turbo), while its 128-NFE quality rivals closed-source models. 
This unique scalability starkly contrasts with competitors like SD3.5-Large, which collapse at a few steps, and FLUX.1-Schnell, which degrades at many steps. 
TiM alone shows monotonic quality improvement with NFE. 
This efficiency is further proven on DPGBench, where 8-NFE TiM outperforms 100-NFE baselines like SDXL. 
Finally, TiM demonstrates superior generalization across diverse resolutions and aspect ratios, validating its fundamentally more robust design.

\begin{figure*}[ht]
\begin{center}
\centerline{\includegraphics[width=0.97\linewidth]{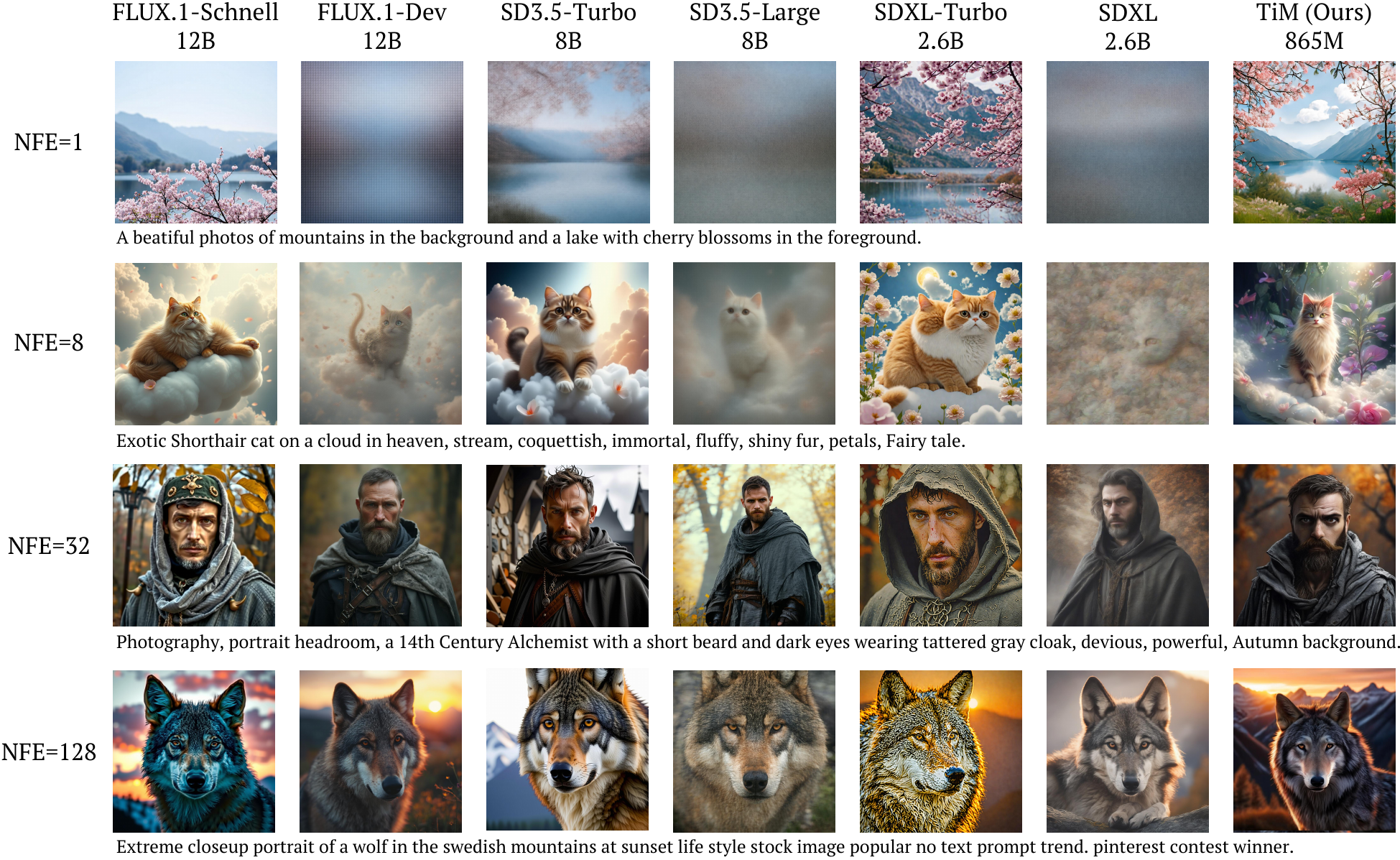}}
\caption{\textbf{Qualitative Analysis between TiM and existing methods under different NFEs.} TiM delivers superior fidelity and text alignment across all NFEs. In contrast, multi-step diffusion and few-step distilled models exhibit pronounced step–quality trade-offs: SDXL, SD3.5-Large, and FLUX.1-Dev fail to generate images at low NFEs, while SDXL-Turbo, SD3.5-Turbo, and FLUX.1-Schnell produce over-saturated outputs at high NFEs.}
\vspace{-8mm}
\label{fig:nfe_comparison}
\end{center}
\end{figure*}

\subsection{Ablation Studies}
We conduct a series of ablation studies to validate our design choices, building from a standard diffusion baseline (SiT-B/4~\cite{ma2024sit} here) to our final TiM configuration. 
We use a 131M parameter model trained on ImageNet-$256\times256$ for 80 epochs and report FID at 1, 8, and 50 NFEs, corresponding to single-step, few-step, and multi-step generation\footnote{1-NFE: single sampling step; 8-NFE: 4 sampling steps with CFG; 50-NFE: 25 sampling steps with CFG.}. 
The results are summarized in Table~\ref{tab:architecture}.

\noindent\textbf{Transition Objective.}
As shown in Table~\ref{tab:architecture} (a vs. c), switching from the standard SiT objective to our TiM objective delivers a dramatic improvement in few-step performance, \textit{reducing the 1-NFE FID by over 6$\times$} (309.5 $\to$ 49.91) while maintaining strong many-step quality. 
This confirms that learning arbitrary transitions is critical for few-step generation. 
Furthermore, our proposed DDE method (c) achieves this performance while being far more scalable than JVP (b), making large-scale training practical.

\noindent\textbf{Architectural Contributions.}
We next analyze the impact of our architectural innovations on top of a vanilla TiM baseline (d). 
Both the \textit{Decoupled Time Embedding} (e) and \textit{Interval-Aware Attention} (f) individually provide substantial gains across all sampling steps. 
Crucially, combining them (g) yields the best performance, lowering the 8-NFE FID from 33.08 to 29.21. 
This demonstrates that enabling the model to explicitly reason about both absolute time and the transition interval is \textbf{complementary and essential} for optimal performance.

\noindent\textbf{Training Strategy.}
Building on our best architecture (g), we apply our proposed \textit{interval weighting} scheme. 
This final step provides a consistent boost across the board (h), further refining the model and achieving our best FID scores of 47.46 / 24.62 / 17.10. 
\section{Conclusion and Limitations}

This paper introduces the Transition Models (TiM), a novel generative model that learns to navigate the entire generative trajectory with unprecedented flexibility. 
The success of our compact 865M model in outperforming multi-billion parameter giants is not just a new state-of-the-art; it is a testament to a more efficient and powerful paradigm. 
By achieving monotonic quality improvement from one step to many, and scaling to ultra-high resolutions, TiM demonstrates that a unified model is not only possible but superior. 
We believe this work paves the way for a new generation of foundation models that are at once efficient, scalable, and promising in their creative potential.

\paragraph{Limitations.} Although TiM delivers a significant contribution to the fundamental generative models, ensuring content safety and controllability remains an open challenge, and model fidelity can degrade in scenarios requiring fine-grained detail, such as rendering text and hands. 
We also observe occasional artifacts at high resolutions (e.g., $3072\times4096$), likely attributable to biases in the underlying autoencoder.

{
    \small
    \bibliographystyle{ieeenat_fullname}
    \bibliography{egbib}
}

\clearpage

\onecolumn
\appendix
\section*{Appendix}

We include additional derivations, experimental details, and results in the appendix. 
\begin{itemize}
    \item In \cref{appx:tim_framework}, we provide a detailed formula derivation of the State Transition Identity, training, and sampling algorithms.
    \item In \cref{appx:connections}, we discuss TiM's relationships with existing methods, including diffusion models, consistency models, and other training approaches.
    \item In \cref{appx:implementation}, we provide the implementation details of text-to-image generation, including native-resolution training, resolution-dependent timestep shifting, model-guidance training, and from-scratch training. 
    \item In \cref{appx:additional_results}, we provide additional ablation results and results on class-guided image generation.
    \item In \cref{appx:qualitative}, we provide more qualitative results of TiM. 
\end{itemize}

\section{Transition Model Framework}~\label{appx:tim_framework}

In this section, we first provide the derivation of the TiM identity equation \cref{eq:tim_identity}. Then we provide the training and sampling algorithms. Finally, we provide a systematic analysis of the connections with other existing methods.

\subsection{TiM Identity Equation Derivation}~\label{appx:tim_derivation}

We demonstrate the derivation from \cref{eq:tim_differentiate} to the TiM identity equation \cref{eq:tim_identity}. We start from the detailed expansion of each term of \cref{eq:tim_differentiate}. Firstly, we have:
\begin{align}
    & \x_t=\alpha_t\x + \sigma_t\veps \label{eq:x_t}, \\
    & \frac{\dm\x_t}{\dt} = \frac{\dm\alpha_t}{\dt}\x + \frac{\dm\sigma_t}{\dt}\veps, \label{eq:dx_dt}
\end{align}
where $\frac{\dm\x_t}{\dt}$ is the PF-ODE of diffusion and \cref{eq:dx_dt} has already been proved in previous works~\citep{lipman2022flowmatching,song2020sde,lu2024scm,sun2025ucgm}.
For $A_{t,r}=\frac{\alpha_r\hat{\sigma}_t-\sigma_r\hat{\alpha}_t}{\hat{\sigma}_t\alpha_t-\hat{\alpha}_t\sigma_t}$, $B_{t,r}=\frac{\sigma_r\alpha_t-\alpha_r\sigma_t}{\hat{\sigma}_t\alpha_t-\hat{\alpha}_t\sigma_t}$, we have:
\begin{align}
    & \frac{\dm A_{t,r}}{\dt} = 
    \frac{\dm A_{t,r}}{\dm\alpha_t}\cdot\frac{\dm\alpha_t}{\dt} + 
    \frac{\dm A_{t,r}}{\dm\sigma_t}\cdot\frac{\dm\sigma_t}{\dt} +
    \frac{\dm A_{t,r}}{\dm\hat{\alpha}_t}\cdot\frac{\dm\hat{\alpha}_t}{\dt} +
    \frac{\dm A_{t,r}}{\dm\hat{\sigma}_t}\cdot\frac{\dm\hat{\sigma}_t}{\dt} 
    \label{eq:dA_dt}
    \\
    & \frac{\dm B_{t,r}}{\dt} = 
    \frac{\dm B_{t,r}}{\dm\alpha_t}\cdot\frac{\dm\alpha_t}{\dt} + 
    \frac{\dm B_{t,r}}{\dm\sigma_t}\cdot\frac{\dm\sigma_t}{\dt} +
    \frac{\dm B_{t,r}}{\dm\hat{\alpha}_t}\cdot\frac{\dm\hat{\alpha}_t}{\dt} +
    \frac{\dm B_{t,r}}{\dm\hat{\sigma}_t}\cdot\frac{\dm\hat{\sigma}_t}{\dt} 
    \label{eq:dB_dt}
\end{align}

We use $C_{t,r}=\hat{\sigma}_t\alpha_t-\hat{\alpha}_t\sigma_t$ for simplicity, which is the denominator of $A_{t,r}$ and $B_{t,r}$. For \cref{eq:dA_dt} and \cref{eq:dB_dt}, each term is calculated as:
\begin{equation}
    \begin{cases}
        & \frac{\dm A_{t,r}}{\dm\alpha_t}=\frac{(\sigma_r\hat{\alpha}_t-\alpha_r\hat{\sigma}_t)\hat{\sigma}_t}{(\hat{\sigma}_t\alpha_t-\hat{\alpha}_t\sigma_t)^2}=-\hat{\sigma}_t\frac{A_{t,r}}{C_{t,r}} \\
        & \frac{\dm A_{t,r}}{\dm\sigma_t}=\frac{(\alpha_r\hat{\sigma}_t-\sigma_r\hat{\alpha}_t)\hat{\alpha}_t}{(\hat{\sigma}_t\alpha_t-\hat{\alpha}_t\sigma_t)^2}=\hat{\alpha}_t\frac{A_{t,r}}{C_{t,r}} \\
        & \frac{\dm A_{t,r}}{\dm\hat{\alpha}_t}=\frac{(\alpha_r\sigma_t-\sigma_r\alpha_t)\sigma_t}{(\hat{\sigma}_t\alpha_t-\hat{\alpha}_t\sigma_t)^2}=-\hat{\sigma}_t\frac{B_{t,r}}{C_{t,r}}\\
        & \frac{\dm A_{t,r}}{\dm\hat{\sigma}_t}=\frac{(\alpha_t\sigma_r-\alpha_r\sigma_t)\alpha_t}{(\hat{\sigma}_t\alpha_t-\hat{\alpha}_t\sigma_t)^2}=\hat{\alpha}_t\frac{B_{t,r}}{C_{t,r}}\\
    \end{cases}
    ;\quad\quad
    \begin{cases}
        & \frac{\dm B_{t,r}}{\dm\alpha_t}=\frac{(\alpha_r\hat{\sigma}_t-\sigma_r\hat{\alpha}_t)\sigma_t}{(\hat{\sigma}_t\alpha_t-\hat{\alpha}_t\sigma_t)^2}=\sigma_t\frac{A_{t,r}}{C_{t,r}} \\
        & \frac{\dm B_{t,r}}{\dm\sigma_t}=\frac{(\sigma_r\hat{\alpha}_t-\alpha_r\hat{\sigma}_t)\alpha_t}{(\hat{\sigma}_t\alpha_t-\hat{\alpha_t}\sigma_t)^2}=-\alpha_t\frac{A_{t,r}}{C_{t,r}} \\
        & \frac{\dm B_{t,r}}{\dm\hat{\alpha}_t}=\frac{(\sigma_r\alpha_t-\alpha_r\sigma_t)\sigma_t}{(\hat{\sigma}_t\alpha_t-\hat{\alpha_t}\sigma_t)^2}=\sigma_t\frac{B_{t,r}}{C_{t,r}} \\
        & \frac{\dm B_{t,r}}{\dm\hat{\sigma}_t}=\frac{(\alpha_r\sigma_t-\sigma_r\alpha_t)\alpha_t}{(\hat{\sigma}_t\alpha_t-\hat{\alpha_t}\sigma_t)^2}=-\alpha_t\frac{B_{t,r}}{C_{t,r}}
    \end{cases}.
    \label{eq:A_B_terms}
\end{equation}

Substituting \cref{eq:A_B_terms} into \cref{eq:dA_dt} and \cref{eq:dB_dt}, we have:
\begin{align}
    & \frac{\dm A_{t,r}}{\dt} = 
    - \hat{\sigma}_t\frac{A_{t,r}}{C_{t,r}}\cdot\frac{\dm\alpha_t}{\dt} 
    + \hat{\alpha}_t\frac{A_{t,r}}{C_{t,r}}\cdot\frac{\dm\sigma_t}{\dt} 
    - \hat{\sigma}_t\frac{B_{t,r}}{C_{t,r}}\cdot\frac{\dm\hat{\alpha}_t}{\dt} 
    + \hat{\alpha}_t\frac{B_{t,r}}{C_{t,r}}\cdot\frac{\dm\hat{\sigma}_t}{\dt},
    \label{eq:dA_dt_plus}
    \\
    & \frac{\dm B_{t,r}}{\dt} = 
      \sigma_t\frac{A_{t,r}}{C_{t,r}}\cdot\frac{\dm\alpha_t}{\dt} 
    - \alpha_t\frac{A_{t,r}}{C_{t,r}}\cdot\frac{\dm\sigma_t}{\dt} 
    + \sigma_t\frac{B_{t,r}}{C_{t,r}}\cdot\frac{\dm\hat{\alpha}_t}{\dt} 
    - \alpha_t\frac{B_{t,r}}{C_{t,r}}\cdot\frac{\dm\hat{\sigma}_t}{\dt}.
    \label{eq:dB_dt_plus}
\end{align}
There exists some symmetry between the above two equations, which is the key to our TiM identity. 
Combining \cref{eq:x_t,eq:dx_dt,eq:dA_dt_plus}, we have:
\begin{equation}
    \x_t\frac{\dm A_{t,r}}{\dt} + A_{t,r}\frac{\dm\x_t}{\dt} = (A_{t,r}\frac{\dm\alpha_t}{\dt}+\alpha_t\frac{\dm A_{t,r}}{\dt})\x + (A_{t,r}\frac{\dm\sigma_t}{\dt}+\sigma_t\frac{\dm A_{t,r}}{\dt})\veps.
    \label{eq:x_t_A_tr}
\end{equation}

The coefficient of $\x$ in the above equation can be decomposed as: 
\begin{equation}
    \begin{aligned}
        & A_{t,r}\frac{\dm\alpha_t}{\dt}+\alpha_t\frac{\dm A_{t,r}}{\dt} 
        \\
        = & 
        (A_{t,r}+\alpha_t\frac{\dm A_{t,r}}{\dm\alpha_t})\frac{\dm\alpha_t}{\dt} 
        + \alpha_t\frac{\dm A_{t,r}}{\dm\sigma_t}\cdot\frac{\dm\sigma_t}{\dt} 
        + \alpha_t\frac{\dm A_{t,r}}{\dm\hat{\alpha}_t}\cdot\frac{\dm\hat{\alpha}_t}{\dt} 
        + \alpha_t\frac{\dm A_{t,r}}{\dm\hat{\sigma}_t}\cdot\frac{\dm\hat{\sigma}_t}{\dt} 
        \\
        = & 
        - \hat{\alpha}_t\sigma_t\frac{A_{t,r}}{C_{t,r}}\cdot\frac{\dm\alpha_t}{\dt} 
        + \hat{\alpha_t}\alpha_t\frac{A_{t,r}}{C_{t,r}}\cdot\frac{\dm\sigma_t}{\dt}
        - \hat{\sigma}_t\alpha_t\frac{B_{t,r}}{C_{t,r}}\cdot\frac{\dm\hat{\alpha}_t}{\dt}
        + \hat{\alpha}_t\alpha_t\frac{B_{t,r}}{C_{t,r}}\cdot\frac{\dm\hat{\sigma}_t}{\dt}
        \\
        = &
        - \hat{\alpha}_t\frac{\dm B_{t,r}}{\dt} + (\hat{\alpha}_t\sigma_t
        - \hat{\sigma}_t\alpha_t)\frac{B_{t,r}}{C_{t,r}}\cdot\frac{\hat{\alpha}_t}{\dt}
        \\ 
        = &
        - \hat{\alpha}_t\frac{\dm B_{t,r}}{\dt} - B_{t,r}\frac{\dm\hat{\alpha}_t}{\dt}.
    \end{aligned}
    \label{eq:coefficient_x}
\end{equation}

Similarly, the coefficient of $\veps$ in the \cref{eq:x_t_A_tr} can be decomposed as:
\begin{equation}
    \begin{aligned}
        & A_{t,r}\frac{\dm\sigma_t}{\dt}+\sigma_t\frac{\dm A_{t,r}}{\dt} 
        \\
        = &
        \sigma_t\frac{\dm A_{t,r}}{\dm\alpha_t}\cdot\frac{\dm\alpha_t}{\dt} 
        + (A_{t,r}+\sigma_t\frac{\dm A_{t,r}}{\dm\sigma_t})\frac{\dm A_{t,r}}{\dm\sigma_t}\cdot\frac{\dm\sigma_t}{\dt} 
        + \sigma_t\frac{\dm A_{t,r}}{\dm\hat{\alpha}_t}\cdot\frac{\dm\hat{\alpha}_t}{\dt} 
        + \sigma_t\frac{\dm A_{t,r}}{\dm\hat{\sigma}_t}\cdot\frac{\dm\hat{\sigma}_t}{\dt} 
        \\
        = &
        - \hat{\sigma}_t\sigma_t\frac{A_{t,r}}{C_{t,r}}\cdot\frac{\dm\hat{\sigma}_t}{\dt}
        + \hat{\sigma}_t\alpha_t\frac{A_{t,r}}{C_{t,r}}\cdot\frac{\dm\hat{\sigma}_t}{\dt}
        - \hat{\sigma}_t\sigma_t\frac{B_{t,r}}{C_{t,r}}\cdot\frac{\dm\hat{\sigma}_t}{\dt}
        + \hat{\alpha}_t\sigma_t\frac{B_{t,r}}{C_{t,r}}\cdot\frac{\dm\hat{\sigma}_t}{\dt}
        \\
        =& 
        - \hat{\sigma}_t\frac{\dm B_{t,r}}{\dt} + (\hat{\alpha}_t\sigma_t
        - \hat{\sigma}_t\alpha_t)\frac{B_{t,r}}{\dt}\cdot\frac{\dm\hat{\sigma}_t}{\dt}
        \\
        =&
        - \hat{\sigma}_t\frac{\dm B_{t,r}}{\dt} - B_{t,r}\frac{\dm\hat{\sigma}_t}{\dt}.
    \end{aligned}
    \label{eq:coefficient_veps}
\end{equation}

Substituting \cref{eq:x_t_A_tr,eq:coefficient_x,eq:coefficient_veps} into \cref{eq:tim_differentiate}, we have:
\begin{equation}
    \begin{aligned}
        & \x_t\frac{\dm A_{t,r}}{\dt} + A_{t,r}\frac{\dm\x_t}{\dt} + \f_{\theta,t,r}\frac{\dm B_{t,r}}{\dt} + B_{t,r}\frac{\dm\f_{\theta,t,r}}{\dt}=0.
        \\
        \Rightarrow &
        (- \hat{\alpha}_t\frac{\dm B_{t,r}}{\dt} - B_{t,r}\frac{\dm\hat{\alpha}_t}{\dt})\x +
        (- \hat{\sigma}_t\frac{\dm B_{t,r}}{\dt} - B_{t,r}\frac{\dm\hat{\sigma}_t}{\dt})\veps + \f_{\theta,t,r}\frac{\dm B_{t,r}}{\dt} + B_{t,r}\frac{\dm\f_{\theta,t,r}}{\dt} = 0.
        \\
        \Rightarrow &
        (\hat{\alpha}_t\x+\hat{\sigma}_t\veps-\f_{\theta,t,r})\frac{\dm B_{t,r}}{\dt} +
        B_{t,r}(\frac{\dm\hat{\alpha}_t}{\dt}+\frac{\dm\hat{\sigma}_t}{\dt}-\frac{\dm\f_{\theta,t,r}}{\dt}) = 0
        \\ 
        \Rightarrow &
        (\hat{\alpha}_t\x + \hat{\sigma}_t\veps - \f_{\theta,t,r})\frac{\dm B_{t,r}}{\dt} + B_{t,r}\frac{\dm(\hat{\alpha}_t\x+\hat{\sigma}_t\veps -\f_{\theta,t,r})}{\dt} = 0
        \\
        \Rightarrow & 
        \frac{\dm(B_{t,r}\cdot(\hat{\alpha}_t\x + \hat{\sigma}_t\veps - \f_{\theta,t,r}))}{\dt} = 0.
    \label{eq:tim_differentiate_to_identity}
    \end{aligned}
\end{equation}

This is the TiM identity equation in \cref{eq:tim_identity}, the proof is completed.

\subsection{TiM Training Algorithm}

\begin{algorithm}[h]
\caption{Training Algorithm of Diffusion Transition Models (TiM).}
\label{alg:tim_train}
\begin{algorithmic}
\State \textbf{Input:} dataset $\mathcal{D}$ with standard deviation $\sigmad$, model $\f_\theta$, diffusion parameterization $\{\alpha_t,\sigma_t,\hat{\alpha}_t,\hat{\sigma}_t\}$, weighting $w_{\Delta t}$, learning rate $\eta$, time distribution $\mathcal{T}$, constant $\eps$, constant $c$.
\State \textbf{Init:} $\text{Iters} \leftarrow 0$
\Repeat
\State $\x_\text{d}\sim\mathcal{D},\ \x=\cdata(\x_\text{d}),\ \veps\sim\gN(\mathbf{0},\mathbf{I}),\ r<t\sim\mathcal{T},\ \x_t\leftarrow \alpha_t\x+\sigma_t\veps$ 
\State $B_{t,r}\leftarrow(\sigma_r\alpha_t-\alpha_r\sigma_t)/(\hat{\sigma}_t\alpha_t-\hat{\alpha}_t\sigma_t)$
\State $\frac{\dm\f_{\theta^-}}{\dt}=\frac{1}{2\eps}(\f_{\theta^-}(\x_{t+\eps},t+\eps,r)-\f_{\theta^-}(\x_{t-\eps},t-\eps,r))$ \Comment{DDE Calculation}
\State $\hat{\f}\leftarrow\hat{\alpha}_t\x+\hat{\sigma}_t\veps+(\frac{\dm\hat{\alpha}_t}{\dt}\cdot\x+\frac{\dm\hat{\sigma}_t}{\dt}\cdot\veps-\frac{\dm\f_{\theta^-}}{\dt})\cdot B_{t,r}/\frac{\dm B_{t,r}}{\dt}$ \Comment{TiM Target}
\State $\gL(\theta)\leftarrow\|\f_\theta-\hat{\f}\|_2^2 + L_\text{cos}(\f_{\theta},\hat{\f})$
\State $\gL(\theta)\leftarrow w_{\Delta t}\cdot\gL(\theta)/(\|\gL(\theta)\|+c)$
\State $\theta\leftarrow\theta-\eta\nabla_{\theta}\gL(\theta)$
\State $\text{Iters} \leftarrow \text{Iters} + 1$
\Until{convergence}
\end{algorithmic}
\end{algorithm}

We provide the detailed training algorithm of TiM in \cref{alg:tim_train}. It is noteworthy that the TiM models are entirely trained from scratch. 

\subsection{TiM Sampling Algorithms}

\begin{algorithm}[h]
\caption{Piecewise Sampling Algorithm of Diffusion Transition Models (TiM).}
\label{alg:tim_sample}
\begin{algorithmic}
\State \textbf{Input:} sampling step $N$, miximum timestep $T_\text{max}$, model $\f_{\theta}$, diffusion parameterization $\{\alpha_t,\sigma_t,\hat{\alpha}_t,\hat{\sigma}_t\}$, stochasticity ratio $\rho$. 
\State \textbf{Init:} data $\x_N\sim\gN(\mathbf{0},\mathbf{I})$, timesteps $\gT=\{t_i\}_{i=N}^{0}$ where $t_N=T_\text{max}, t_0=0$
\For{$i = N$ to $1$}
\State $\x_{t_{i+1}}=\frac{\alpha_{t_{i+1}}\hat{\sigma}_{t_i}-\sigma_{t_{i+1}}\hat{\alpha}_{t_i}}{\hat{\sigma}_{t_i}\alpha_{t_i}-\hat{\alpha}_{t_i}\sigma_{t_i}}\x_{t_i} + \frac{\sigma_{t_{i+1}\alpha_{t_i}-\alpha_{t_{i+1}}\sigma_{t_i}}}{\hat{\sigma}_{t_i}\alpha_{t_i}-\hat{\alpha}_{t_i}\sigma_{t_i}}\f(\x_{t_i},t_i,t_{i+1})$
\If{$\rho>0$}:
\State $\hat{\veps}\leftarrow\frac{\alpha_{t_i}}{\hat{\sigma}_{t_i}\alpha_{t_i}\hat{\alpha}_{t_i}\sigma_{t_i}}\f_{\theta}(\x_{t_i},t_i,t_0) - 
\frac{\hat{\alpha}_{t_i}}{\hat{\sigma}_{t_i}\alpha_{t_i}-\hat{\alpha}_{t_i}\sigma_{t_i}}\x_{t_i}$
\State $\veps_i\sim\gN(\mathbf{0},\mathbf{I})$
\State $\dt=t_{i}-t_{i+1}$
\State $\x_{t_{i+1}}\leftarrow\x_{t_{i+1}}-\rho(\alpha_{t_i}\sigma_{t_i}'-\alpha_{t_i}'\sigma_{t_i})\hat{\veps}\dt-\sqrt{2\rho(\alpha_{t_i}\sigma_{t_i}'-\alpha_{t_i}'\sigma_{t_i})}\veps_i\sqrt{\dt}$
\EndIf
\State $\x_{i}=\x_{t_{i+1}}$
\EndFor
\end{algorithmic}
\end{algorithm}

We provide the TiM sampling algorithm in \cref{alg:tim_sample}. For multi-step sampling, we can further incorporate stochasticity into the sampling process for improved diversity. In multi-step scenarios, the TiM sampling is similar to the diffusion sampling process, but with a new condition for the next step. 
Therefore, we can construct a stochastic sampling from the SDE (stochastic differential equation) diffusion process. Given $\x_t=\alpha_t+\sigma_t\veps,  $\citet{song2020sde} has shown that the SDE forward and reverse are:
\begin{equation}
\begin{aligned}
    & \text{forward}:\ \dm\x_t = \mathrm{f}(\x_t,t)+\mathrm{g}(t)\dm\rvw, \\
    & \text{reverse}:\ \dm\x_t = [\mathrm{f}(\x_t,t)-\frac{1}{2}\mathrm{g}(t)^2\nabla_{\x_t}\log p_t(\x_t)]\dt + \mathrm{g}(t)\dm\rvw.
\end{aligned}
\end{equation}

Previous works\citep{song2020sde,karras2022edm,lu2024scm,sun2025ucgm} has provided the explicit form of $\mathrm{f}(\x_t,t)$, $\mathrm{g}(t)$ and $\nabla_{\x_t}\log p_t(\x_t)$:
\begin{equation}
    \mathrm{f}(\x_t,t) = \frac{\dot{\alpha}_t}{\alpha_t}\x_t, \quad
    \mathrm{g}(t) = 2\sigma_t\dot{sigma}_t-2\frac{\dot{\alpha}_t}{\alpha_t}\sigma_t^2, \quad
    \nabla_{\x_t}\log p_t(\x_t) = -\frac{\veps}{\sigma_t},
\end{equation}
where $\dot{\alpha}_t$ and $\dot{\sigma}_t$ represent the derivation of $\alpha_t$ and $\sigma_t$ respectively. 
For PF-ODE, it is defined as: 
\begin{equation}
    \mathbf{v}_t = \frac{\dm\x_t}{\dt}=\mathrm{f}(\x_t,t)-\frac{1}{2}\mathrm{g}(t)^2\nabla_{\x_t}\log p_t(\x_t)=\dot{\alpha}_t\x+\dot{\sigma}_t\veps.
\end{equation}
For reverse-SDE, it is defined as:
\begin{equation}
    \begin{aligned}
        \dm\x_t 
        &= [\mathrm{f}(\x_t,t)-\mathrm{g}(t)^2\nabla_{\x_t}\log p_t(\x_t)]\dt + \mathrm{g}(t)\dm\rvw \\
        &= \underbrace{[\mathrm{f}(\x_t,t)-\frac{1}{2}\mathrm{g}(t)^2\nabla_{\x_t}\log p_t(\x_t)]}_{\text{PF-ODE Term}}\dt \underbrace{- \frac{1}{2}\mathrm{g}(t)^2\nabla_{\x_t}\log p_t(\x_t)\dt + \mathrm{g}(t)\dm\rvw}_{\text{Stochastic Term}} \\
        &= \mathbf{v}_t\dt + [\dot{\sigma}_t-\frac{\dot{\alpha}_t}{\alpha_t}\sigma_t]\veps\dt + \sqrt{2\sigma_t\dot{\sigma}_t-2\frac{\dot{\alpha}_t}{\alpha_t}\sigma_t^2}\dm\rvw.
    \end{aligned}
\end{equation}

In the TiM sampling, we can take the stochastic term in the above equation to enhance diversity. To balance the stochasticity and stability, we incorporate a scaling factor $s(t)=\rho\alpha_t$, leading to a scaled $\tilde{\mathrm{g}}=\rho\alpha_t\mathrm{g}(t)=2\rho(\alpha_t\sigma_t\dot{\sigma}_t-2\dot{\alpha}_t\sigma_t^2)$. Therefore, the stochastic term is: $\rho[\alpha_t\dot{\sigma}_t-\dot{\alpha}_t\sigma_t]\veps\dt + \sqrt{2\rho(\alpha_t\sigma_t\dot{\sigma}_t-2\dot{\alpha}_t\sigma_t^2)}\dm\rvw$.

\section{Connections with Existing Methods}~\label{appx:connections}

In this section, we highlight the connection between TiM and other existing methods. We first demonstrate the properties of TiM compared with diffusion models. Then we demonstrate the connections of TiM with other training strategies.

\begin{table*}[t]
\renewcommand{\arraystretch}{1.2} 
\centering
\begin{adjustbox}{max width=\textwidth}
\begin{tabular}{l|ccccc|cccc}
\toprule[1.2pt]
\multirow{2}{*}{\textbf{Transport}}
& \multicolumn{5}{c|}{\textbf{\textit{Diffusion Parameterization}}}
& \multicolumn{4}{c}{\textbf{\textit{Transition Parameterization}}} \\
\cmidrule(lr){2-6} \cmidrule(lr){7-10} \
& \textbf{$\cnoise(t)=$} 
& \textbf{$\alpha_t=$} 
& \textbf{$\sigma_t=$} 
& \textbf{$\hat{\alpha}_t=$} 
& \textbf{$\hat{\sigma}_t=$}
& \textbf{$\frac{\dm\hat{\alpha}_t}{\dt}=$} 
& \textbf{$\frac{\dm\hat{\sigma}_t}{\dt}=$}
& \textbf{$B_{t,r}=$} 
& \textbf{$\frac{\dm B_{t,r}}{\dt}=$}  \\
\midrule
\textbf{OT-FM~\cite{lipman2022flowmatching,liu2023flow}}
& $t$
& $1-t$
& $t$
& $-1$
& $1$
& $0$
& $0$
& $r-t$
& $-1$
\\ 
\textbf{TrigFlow~\cite{lu2024scm}}
& $t$
& $\cos(t)$
& $\sin(t)$
& $-\sin(t)$
& $\cos(t)$
& $-\cos(t)$ 
& $-\sin(t)$
& $\sin(r-t)$
& $-\cos(r-t)$
\\ 
\textbf{EDM~\cite{karras2022edm,karras2024edm2}}
& $\frac{1}{4}\ln(t)$
& $\frac{1}{t^2+\sigmad^2}$
& $\frac{t}{\sqrt{t^2+\sigmad^2}}$
& $\frac{t}{\sigmad\sqrt{t^2+\sigmad^2}}$
& $-\frac{\sigmad}{\sqrt{t^2+\sigmad^2}}$
& \cref{eq:tim_edm_param_1} & \cref{eq:tim_edm_param_2} & \cref{eq:tim_edm_param_3} & \cref{eq:tim_edm_param_4} \\
\textbf{VP-SDE~\cite{song2020sde,ho2020ddpm}}
& $(T-1)t$
& $\frac{1}{\beta_t^2+1}$
& $\frac{\beta_t}{\sqrt{\beta_t^2+1}}$
& $0$
& $1$
& $0$ 
& $0$
& $\frac{\beta_r-\beta_t}{\sqrt{\beta_r^2+1}}$
& $\frac{-1}{\sqrt{\beta_r^2+1}}\cdot\frac{\dm\beta_t}{\dt}$
\\ 
\textbf{VE-SDE~\cite{song2020sde,song2019ncsn}}
& $\ln(\frac{1}{2}t)$
& $1$
& $t$
& $0$
& $-1$
& $0$
& $0$
& $t-r$
& $1$
\\ 
\bottomrule[1.2pt]
\end{tabular}
\end{adjustbox}
\vspace{-3mm}
\caption{
\textbf{Transition parameterization for different diffusion transports}. For VP-SDE, $T$ is set to $1000$, and $\beta_t=\sqrt{e^{\frac{1}{2}\beta_\text{d} t^2+\beta_\text{min}t}-1}$, where $\beta_\text{d}=19.9$ and $\beta_\text{min}=0.1$ by default. We provide the details of timestep sampling in \cref{tab:time_distribution}. \citet{song2020sde} has shown that VP-SDE is equivalent to DDPM~\cite{ho2020ddpm} while VE-SDE is equivalent to score matching~\cite{song2019ncsn}, so we adopt their notations for uniformity. For EDM, its TiM parameterization is too complex; we provide them in \cref{eq:tim_edm_param_1,eq:tim_edm_param_2,eq:tim_edm_param_3,eq:tim_edm_param_4}.}
\vspace{-3mm}
\label{tab:tim_formulation}
\end{table*}

\begin{table}[htbp]
\renewcommand{\arraystretch}{1.2} 
\centering
\begin{adjustbox}{max width=.8\textwidth}
\begin{tabular}{l|cccc}
\toprule[1.2pt]
Transport
& Noise Level
& Timestep
& Time Range
& Time Scaling
\\
\midrule
\textbf{OT-FM}
& $\ln(\sigma)\sim\gN(P_\text{mean}, P_\text{std}^2)$ 
& $t=\frac{\sigma}{1+\sigma}$ 
& $t\in[0,1]$
& $\cnoise(t)=t$
\\ 
\textbf{Trigflow}
& $\ln(\sigma)\sim\gN(P_\text{mean}, P_\text{std}^2)$ 
& $t=\arctan(\frac{\sigma}{\sigmad})$ 
& $t\in[0,\frac{\pi}{2}]$
& $\cnoise(t)=t$
\\ 
\textbf{EDM}
& $\ln(\sigma)\sim\gN(P_\text{mean}, P_\text{std}^2)$ 
& $t=\sigma$ 
& $t\in[0,+\infty)$
& $\cnoise(t)=\frac{1}{4}\ln(t)$
\\
\textbf{VP}
& $\sigma\sim\gU(\epsilon_t, 1)$ 
& $t=\sigma$ 
& $t\in[\epsilon_t, 1]$
& $\cnoise(t)=(T-1)t$
\\ 
\textbf{VE}
& $\sigma\sim\gU(\epsilon_t, 1)$ 
& $t=\sigma_{\text{max}}(\frac{\sigma_{\text{min}}^2}{\sigma_{\text{max}}^2})^\sigma$
& $t\in[\sigma_{\text{min}}, \sigma_{\text{max}}]$
& $\cnoise(t)=\ln(\frac{1}{2}t)$
\\ 
\bottomrule[1.2pt]
\end{tabular}
\end{adjustbox}
\caption{Time distribution of diffusion diffusion transports.}
\label{tab:time_distribution}
\end{table}

\subsection{Connections with Diffusion Models}

\paragraph{TiM generalizes the standard diffusion models.}
As a complement to \cref{tab:tim_formulation}, we elucidate the time distribution of different diffusion transports in \cref{tab:time_distribution}. Our TiMs share these parameters with diffusion models, but learn a different objective. We show that the TiM training objective ~\cref{eq:tim_objective} generalizes the standard diffusion objective \cref{eq:diffusion_objective}. Specifically, the TiM identity equation reduces to the diffusion identity equation in the limit as $t\rightarrow r$. Recall that $B_{t,r}=\frac{\sigma_r\alpha_t-\alpha_r\sigma_t}{\hat{\sigma}_t\alpha_t-\hat{\alpha}_t\sigma_t}$, the training target of TiM when $t\rightarrow r$ becomes:
\begin{equation}
    \begin{aligned}
        \lim_{t\rightarrow r}\hat{\f} 
        &= \lim_{t\rightarrow r}\left(\hat{\alpha}_t\x+\hat{\sigma}_t\veps+\frac{B_{t,r}}{\frac{\dm B_{t,r}}{\dt}}\left(\frac{\dm\hat{\alpha}_t}{\dt}\cdot\x+\frac{\dm\hat{\sigma}_t}{\dt}\cdot\veps-\frac{\dm\f_{\theta^-,t,r}}{\dt}\right)\right) \\
        &= \hat{\alpha}_t\x+\hat{\sigma}_t\veps+\lim_{t\rightarrow r}\left(\frac{B_{t,r}}{\frac{\dm B_{t,r}}{\dt}}\left(\frac{\dm\hat{\alpha}_t}{\dt}\cdot\x+\frac{\dm\hat{\sigma}_t}{\dt}\cdot\veps-\frac{\dm\f_{\theta^-,t,r}}{\dt}\right)\right) \\
        &= \hat{\alpha}_t\x+\hat{\sigma}_t\veps+\lim_{t\rightarrow r}\left(\frac{\frac{\sigma_r\alpha_t-\alpha_r\sigma_t}{\hat{\sigma}_t\alpha_t-\hat{\alpha}_t\sigma_t}}{\frac{\dm B_{t,r}}{\dt}}\left(\frac{\dm\hat{\alpha}_t}{\dt}\cdot\x+\frac{\dm\hat{\sigma}_t}{\dt}\cdot\veps-\frac{\dm\f_{\theta^-,t,r}}{\dt}\right)\right) \\
        &= \hat{\alpha}_t\x+\hat{\sigma}_t\veps+\lim_{t\rightarrow r}\left(\frac{0}{\frac{\dm B_{t,r}}{\dt}}\left(\frac{\dm\hat{\alpha}_t}{\dt}\cdot\x+\frac{\dm\hat{\sigma}_t}{\dt}\cdot\veps-\frac{\dm\f_{\theta^-,t,r}}{\dt}\right)\right) \\
        &= \hat{\alpha}_t\x+\hat{\sigma}_t\veps.
    \end{aligned}
\end{equation}
The above target is the diffusion target. This target lacks the modeling of state transitions from state to state, thus limiting the arbitrary-step generation capabilities of diffusion models. 

\paragraph{EDM parametrization.} EDM~\citep{karras2022edm,karras2024edm2} parameterizes the diffusion model as: 
\begin{equation}
    \D_{\theta}(\x+t\veps,t)=\frac{\sigmad^2}{t^2+\sigmad^2}(\x+t\veps)+\frac{t\cdot\sigmad}{\sqrt{t^2+\sigmad^2}}\F_{\theta}\left(\frac{\x+t\veps}{\sqrt{t^2+\sigmad^2}},\frac{1}{4}\ln(t)\right).
\end{equation}
It adopts the $\x$-prediction in its training and use time weighting $w(t)=\frac{t^2+\sigmad^2}{t^2\sigmad^2}$, leading to training objective as:
\begin{equation}
    \begin{aligned}
        \gL(\theta)
        &= \frac{t^2+\sigmad^2}{t^2\sigmad^2}\|\D_{\theta}(\x+t\veps,t)-\x\|_2^2 \\
        &= \frac{t^2+\sigmad^2}{t^2\sigmad^2}\left\|\frac{\sigmad^2}{t^2+\sigmad^2}(\x+t\veps)+\frac{t\cdot\sigmad}{\sqrt{t^2+\sigmad^2}}\F_{\theta}\left(\frac{\x+t\veps}{\sqrt{t^2+\sigmad^2}},\frac{1}{4}\ln(t)\right)-\x\right\|_2^2 \\
        &= \left\|\F_{\theta}\left(\frac{\x+t\veps}{\sqrt{t^2+\sigmad^2}},\frac{1}{4}\ln(t)\right)-(\frac{t}{\sigmad\sqrt{t^2+\sigmad^2}}\x-\frac{\sigmad}{\sqrt{t^2+\sigmad^2}}\veps)\right\|_2^2 \\
    \end{aligned}
\end{equation}

Therefore, let $\cnoise(t)=\frac{1}{4}\ln(t)$, the original EDM parameterization can be unified into our parameterization with the following coefficients:
\begin{equation}
    \alpha_t=\frac{1}{\sqrt{t^2+\sigmad^2}},\quad
    \sigma_t=\frac{t}{\sqrt{t^2+\sigmad^2}},\quad
    \hat{\alpha}_t=\frac{t}{\sigmad\sqrt{t^2+\sigmad^2}}\quad 
    \hat{\sigma}_t=-\frac{\sigmad}{\sqrt{t^2+\sigmad^2}}
\end{equation}
Therefore, the TiM parameterization is defined as:
\begin{align}
    \frac{\dm\hat{\alpha}_t}{\dt} 
    &= -\frac{t^2}{\sigmad(t^2+\sigmad^2)^{\frac{3}{2}}} + \frac{1}{\sigmad\sqrt{t^2+\sigmad^2}}, \label{eq:tim_edm_param_1}\\
    \frac{\dm\hat{\sigma}_t}{\dt} 
    &= \frac{t\sigmad}{(t^2+\sigmad^2)^{\frac{3}{2}}}, \label{eq:tim_edm_param_2}\\
    B_{t,r} 
    &= \frac{(t-r)\sigmad^2\sqrt{t^2+\sigmad^2}}{(t^2+\sigmad^3)\sqrt{r^2+\sigmad^2}}, \label{eq:tim_edm_param_3}\\
    \frac{\dm B_{t,r}}{\dt} 
    &= \sigmad^2\frac{t(t-r)(t^2+\sigmad^3)-2t(t-r)(t^2+\sigmad^2)+(t^2+\sigmad^2)(t^2+\sigmad^3)}{(t^2+\sigmad^3)^2\sqrt{t^2+\sigmad^2}\sqrt{r^2+\sigmad^2}}. \label{eq:tim_edm_param_4}
\end{align}

\subsection{Connections to Other training Methods}

In this section, we discuss the connections of TiM with other training strategies, including continuous-time consistency models~\citep{song2023cm,lu2024scm},  consistency trajectory models~\citep{kim2023ctm}, phased consistency models~\citep{wang2025pcm}, Shortcut models~\citep{frans2024shortcut}, and MeanFlow models~\citep{geng2025meanflow,peng2025facm}. 

\paragraph{Continuous-time consistency models.} The TiM objective \cref{eq:tim_objective} generalizes the continuous-time consistency models. Specifically, the CTM objective reduces to the continuous-time CM objective when $r=0$.
For TiM, let $r=0$ and $d(\x,\y)=\|\x-\y\|_2^2$, the training objective becomes:
\begin{equation}
    \begin{aligned}
        &\nabla_\theta\E_{\x,\veps,t}\left[\left\|\f_\theta(\x_t,t,0)-(\hat{\alpha}_t\x+\hat{\sigma}_t\veps+\frac{B_{t,0}}{\frac{\dm B_{t,0}}{\dt}}\left(\frac{\dm\hat{\alpha}_t}{\dt}\cdot\x+\frac{\dm\hat{\sigma}_t}{\dt}\cdot\veps-\frac{\dm\f_{\theta^-,t,0}}{\dt}\right)\right\|_2^2\right]
        \\
        =
        &\E_{\x,\veps,t}\left[[\nabla_\theta\f_{\theta,t,0}]^T\left(\f_{\theta^-,t,0}-\hat{\alpha}_t\x-\hat{\sigma}_t\veps+\frac{B_{t,0}}{\frac{\dm B_{t,0}}{\dt}}\left(\frac{\dm\f_{\theta^-,t,0}}{\dt}-\frac{\dm\hat{\alpha}_t}{\dt}\cdot\x-\frac{\dm\hat{\sigma}_t}{\dt}\cdot\veps\right)\right)\right].
    \end{aligned}
    \label{eq:gradient_tim_r0}
\end{equation}

Continuous-time consistency models~\citep{song2023cm,song2023icm,lu2024scm} are trained to map the noisy input $\x_t$ directly to the clean data $\x$ in one or a few steps. Given model $F_\theta$, the consistency models are formulated as:
\begin{equation}
    \D_\theta(\x_t,t)=\cskip(t)\x_t+\cout(t)\F_\theta(\cin(t)\x_t,\cnoise(t)).
\end{equation}
Using the parameters $\alpha_t$, $\sigma_t$, $\hat{\alpha}_t$, and $\hat{\sigma}_t$, consistency parameterization corresponds to the transition from $\x_t$ to $\x_0$: 
\begin{equation}
    \D_\theta(\x_t,t) = \frac{\hat{\sigma}_t\x_t-\sigma_t\F_{\theta}(\cin(t)\x_t,\cnoise(t))}{\hat{\sigma}_t\alpha_t-\hat{\alpha}_t\sigma_t},
\end{equation}
where $\frac{\hat{\sigma}_t}{\hat{\sigma}_t\alpha_t-\hat{\alpha}_t\sigma_t}=A_{t,0}$ and $-\frac{\sigma_t}{\hat{\sigma}_t\alpha_t-\hat{\alpha}_t\sigma_t}=B_{t,0}$ correspond to TiM parameterizations. 

When using loss function $d(\x,\y)=\|\x-\y\|_2^2$, \cite{song2023cm} show that the gradient of continuous-time consistency models is:
\begin{equation}
\begin{aligned}
    & \nabla_\theta\E_{\x_t,t}\left[\D_\theta^T(\x_t,t)\frac{\dm\D_{\theta^-}(\x_t,t)}{\dt} \right] 
    \\
    = & \nabla_\theta\E_{\x_t,t}\left[[B_{t,0}\f_{\theta}^{\text{cm}}(\x_t,t)]^T\frac{\dm\D_{\theta^-}(\x_t,t)}{\dt} \right] 
    \\
    = & \E_{\x_t,t}\left[ B_{t,0}[\nabla_\theta\f_{\theta}^{\text{cm}}(\x_t,t)]^T\frac{\dm\D_{\theta^-}(\x_t,t)}{\dt} \right]
    \\
    = & 
    \E_{\x_t,t}\left[ B_{t,0} \nabla_\theta[\f_{\theta}^{\text{cm}}(\x_t,t)]^T \left(\frac{\dm A_{t,0}}{\dt}\x_t+A_{t,0}\frac{\x_t}{\dt}+\frac{\dm B_{t,0}}{\dt}\f_{\theta^-}^{\text{cm}}+B_{t,0}\frac{\dm\f_{\theta^-}^{\text{cm}}}{\dt}\right) \right],
\end{aligned}
\label{eq:gradient_consistency}
\end{equation}
where $\f_\theta^{\text{cm}}(\x_t,t)=\F_{\theta}(\cin(t)\x_t,\cnoise(t)))$ represents the network in consistency models.
As $\x_t=\alpha_t\x+\sigma_t\veps, \frac{\dm\x_t}{\dt}=\frac{\dm\alpha_t}{\dt}\x+\frac{\sigma_t}{\dt}\veps$, we have:
\begin{equation}
\begin{aligned}
    & \frac{\dm A_{t,0}}{\dt}\x_t+A_{t,0}\frac{\x_t}{\dt}+\frac{\dm B_{t,0}}{\dt}\f_{\theta^-}^{\text{cm}}+B_{t,0}\frac{\dm\f_{\theta^-}^{\text{cm}}}{\dt}
    \\ 
    = &
    \frac{\dm A_{t,0}}{\dt}(\alpha_t\x+\sigma_t\veps)+A_{t,0}(\frac{\dm\alpha_t}{\dt}\x+\frac{\dm\sigma_t}{\dt}z)+\frac{\dm B_{t,0}}{\dt}\f_{\theta^-}^{\text{cm}}+B_{t,0}\frac{\dm\f_{\theta^-}^{\text{cm}}}{\dt}
    \\
    =& 
    (\alpha_t\frac{\dm A_{t,0}}{\dt}+A_{t,0}\frac{\dm\alpha_t}{\dt})\x + (\sigma_t\frac{\dm A_{t,0}}{\dt}+A_{t,0}\frac{\dm\sigma_t}{\dt})\z + \frac{\dm B_{t,0}}{\dt}\f_{\theta^-}^{\text{cm}} + B_{t,0}\frac{\dm\f_{\theta^-}^{\text{cm}}}{\dt}.
\end{aligned}
\end{equation}

Based on \cref{eq:coefficient_x,eq:coefficient_veps}, we have:
\begin{equation}
    \alpha_t\frac{\dm A_{t,0}}{\dt}+A_{t,0}\frac{\dm\alpha_t}{\dt} = -\hat{\alpha}_t\frac{\dm B_{t,0}}{\dt} - B_{t,0}\frac{\dm\hat{\alpha}_t}{\dt}, 
    \quad
    \sigma_t\frac{\dm A_{t,0}}{\dt}+A_{t,0}\frac{\dm\sigma_t}{\dt} = - \hat{\sigma}_t\frac{\dm B_{t,0}}{\dt} - B_{t,0}\frac{\dm\hat{\sigma}_t}{\dt}.
\end{equation}
Therefore, we have:
\begin{equation}
\begin{aligned}
    & \frac{\dm A_{t,0}}{\dt}\x_t+A_{t,0}\frac{\x_t}{\dt}+\frac{\dm B_{t,0}}{\dt}\f_{\theta^-}^{\text{cm}}+B_{t,0}\frac{\dm\f_{\theta^-}^{\text{cm}}}{\dt}
    \\ 
    =& 
    B_{t,0}(\frac{\dm\f_{\theta^-}^{\text{cm}}}{\dt}-\frac{\dm\hat{\alpha}_t}{\dt}\x-\frac{\dm\hat{\sigma}_t}{\dt}) + \frac{\dm B_{t,0}}{\dt}(\f_{\theta^-}^{\text{cm}}-\hat{\alpha}_t\x-\hat{\sigma}_t\z).
\end{aligned}
\end{equation}

Substituting the above equation into \cref{eq:gradient_consistency}, the gradient of continuous-time consistency models is:

\begin{equation}
\begin{aligned}
    & \nabla_\theta\E_{\x_t,t}\left[\D_\theta^T(\x_t,t)\frac{\dm\D_{\theta^-}(\x_t,t)}{\dt} \right] 
    \\
    = & 
    \E_{\x_t,t}\left[ B_{t,0} [\nabla_\theta\f_{\theta}]^T\left( B_{t,0}(\frac{\dm\f_{\theta^-}^{\text{cm}}}{\dt}-\frac{\dm\hat{\alpha}_t}{\dt}\x-\frac{\dm\hat{\sigma}_t}{\dt}) + \frac{\dm B_{t,0}}{\dt}(\f_{\theta^-}^{\text{cm}}-\hat{\alpha}_t\x-\hat{\sigma}_t\z) \right) \right],
    \\
    = &
    \E_{\x_t,t}\left[ (B_{t,0}\frac{\dm B_{t,0}}{\dt}) [\nabla_\theta\f_{\theta}]^T\left(\f_{\theta^-}^{\text{cm}}-\hat{\alpha}_t\x-\hat{\sigma}_t\z + \frac{B_{t,0}}{\frac{\dm B_{t,0}}{\dt}}(\frac{\dm\f_{\theta^-}^{\text{cm}}}{\dt}-\frac{\dm\hat{\alpha}_t}{\dt}\x-\frac{\dm\hat{\sigma}_t}{\dt}) \right) \right],
\end{aligned}
\label{eq:gradient_consistency_final}
\end{equation}

Note that TiM network $\f_\theta(\x_t,t,0)$ corresponds to the network $\f_\theta^{\text{cm}}(\x_t,t)$ in consistency models. The only difference between \cref{eq:gradient_consistency_final} and \cref{eq:gradient_tim_r0} is a term $(B_{t,0}\frac{\dm B_{t,0}}{\dt})$, which can be bridged by a weighting function.

\paragraph{Consistency trajectory models, phased consistency models, and shortcut models.} These models learn to transition from one state to another state in a discrete manner, while our TiM generalizes this to the continuous-time domain. The core of our method is the TiM identity equation \cref{eq:tim_identity}, which determines the function for the transition between two arbitrary states. 
For consistency trajectory models (CTM)~\citep{kim2023ctm} and phased consistency models (PCM)~\citep{wang2025pcm}, they targets at intermediate state $\x_r$, where $0\leqslant r\leqslant t_{n-1}$, thus leading to the identity equation:
\begin{equation}
    \Psi(\x_{t_n},\f_\theta(\x_{t_n},t_n,r),r) = \Psi(\x_{t_{n-1}}, \f_\theta(\x_{t_{n-1}},t_{n-1},r),
    \label{eq:ctm_pcm_identity}
\end{equation}
where $0<t_1<\cdots<t_n<\cdots<t_N=T$ represents the discrete timesteps, and $\Psi$ is an ODE solver to obtain the state at timemstep $r$. It is noteworthy that PCM splits the entire trajectory into several segments and learns this identity on each segment independently. 

Shortcut models~\citep{frans2024shortcut} adopts the OT-flow-matching~\citep{lipman2022flowmatching} as the transport, the ODE solver is: 
$\Psi(\x_t,\f_\theta(\x_t,t,r),r)=\x_t-(t-r)\f_\theta(\x_t,t,r)$. 
The original identity equation is: $(t-r)\f_\theta(\x_t,t,r)=(t-s)\f_\theta(\x_t,t,s)+(s-r)\f_\theta(\x_s,s,r)$. 
This identity equation of shortcut models can be rearranged as:
\begin{equation}
    \begin{aligned}
        & (t-r)\f_\theta(\x_t,t,r)=(t-s)\f_\theta(\x_t,t,s)+(s-r)\f_\theta(\x_s,s,r)
        \\
        \Longrightarrow &
        \x_t-(t-r)\f_\theta(\x_t,t,r)=\x_t-(t-s)\f_\theta(\x_t,t,s)-(s-r)\f_\theta(\x_s,s,r)
        \\
        \Longrightarrow &
        \x_t-(t-r)\f_\theta(\x_t,t,r)=\x_s-(s-r)\f_\theta(\x_s,s,r)
        \\
        \Longrightarrow &
        \Psi(\x_t,\f_\theta(\x_t,t,r),r)=\Psi(\x_s,\f_\theta(\x_s,s,r),r),
    \end{aligned}
    \label{eq:shortcut_identity}
\end{equation}
where $s=\frac{t+r}{2}$. Based on \cref{eq:ctm_pcm_identity} and \cref{eq:shortcut_identity}, when $t_{n-1}=\frac{t_n+r}{2}$, CTMs are equivalent to shortcut models.

\paragraph{MeanFlow models.} We show that the TiM \cref{eq:tim_identity} generalizes the MeanFlow~\citep{geng2025meanflow}. In particular, the training objective of TiM reduces to the MeanFlow objective in the OT-FM~\citep{lipman2022flowmatching} transport setting.

As in \cref{tab:tim_formulation}, OT-FM uses the parameterization $\{\alpha_t=1-t,\sigma_t=t,\hat{\alpha}_t=-1,\hat{\sigma}_t=1\}$, leading to the TiM parameterization $\{B_{t,r}=r-t, \frac{\dm B_{t,r}}{\dt}=-1\}$. Therefore, the TiM training objective becomes:
\begin{equation}
    \E_{\x,\veps,t}\left[ d\left( \f_\theta(\x_t,t,r) - \left( \z-\x -(t-r)\frac{\dm\f_{\theta^-,t,r}}{\dt} \right) \right) \right].
\end{equation}

This corresponds to the training objective of MeanFlow.

\section{Implementation Details}~\label{appx:implementation}

\subsection{Model Architecture}

\begin{figure*}[t]
\begin{center}
\centering
\begin{minipage}{0.62 \textwidth}
\centering
\centerline{\includegraphics[width=\linewidth]{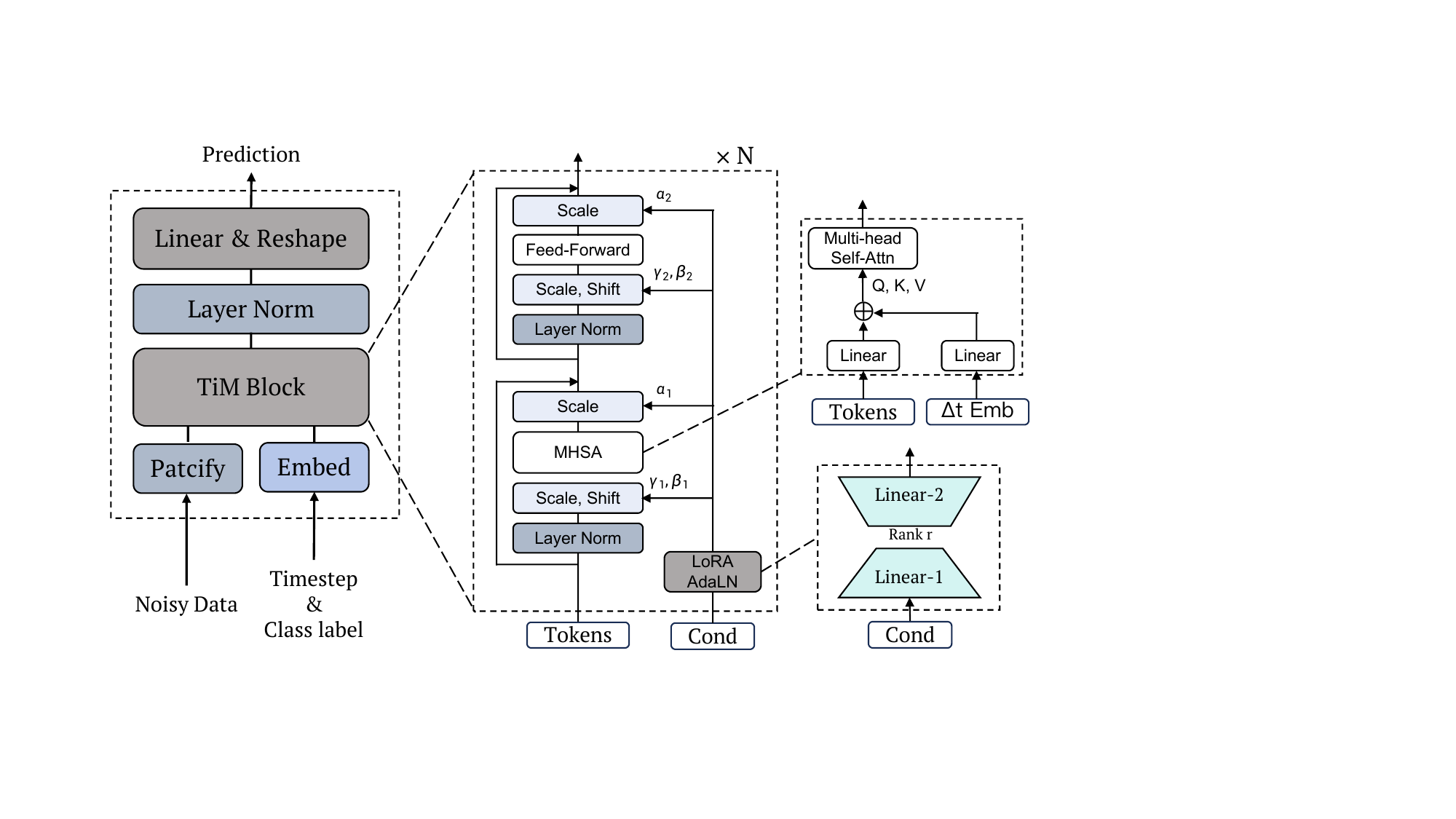}}
\caption{\textbf{TiM Model Architecture.} }
\label{fig:model_architecture}
\end{minipage}
\hfill
\begin{minipage}{0.35\textwidth}
\centering
\centerline{\includegraphics[width=\linewidth]{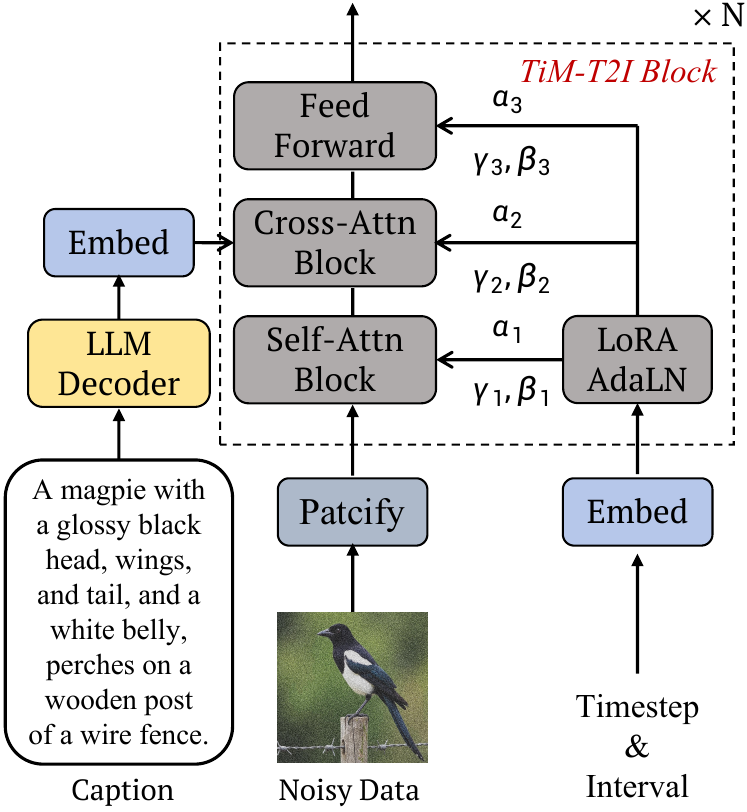}}
\caption{\textbf{TiM T2I block.}}
\label{fig:t2i_block}
\end{minipage}
\end{center}
\end{figure*}

We illustrate the model architecture in \cref{fig:model_architecture}. As we incorporate decoupled time embedding and interval-aware attention designs into DiT architecture, we use LoRA-AdaLN to avoid increasing model size. Specifically, given  attention hidden size $D$, LoRA rank is set as $r=\frac{1}{3}D$, such as $D=1152$ and $r=384$ in XL-models. For text-to-image generation, we incorporate caption features through \texttt{CrossAttention} mechanism, as in \cref{fig:t2i_block}.

\subsection{Text-to-Image Training Details}\label{appx:training_details}

\paragraph{Native-Resolution Training.} 
We adopt the VAE-specific native resolution training for text-to-image generation. As we use DC-AE~\cite{chen2024dc-ae} with $32$ downsampling scale, an image with shape $H\times W$ is resized to shape $(32\cdot\lfloor\frac{H}{32}\rfloor)\times(32\cdot\lfloor\frac{W}{32}\rfloor)$. For example, an image with shape $1025\times513$ is resized to $1024\times512$, , preserving resolution and aspect ratio as much as possible. Images of the same size are grouped into resolution buckets for batching.

We set the base batch size as $16$ on a single GPU for $1024\times1024$ resolution bucket, then for $H\times W$ resolution bucket, the minimal batch size is $B=\lfloor\frac{16\times H\times W}{1024\times1024}\rceil$. For instance, the $512\times512$ resolution bucket holds the minimal batch size as $B=64$, while the $2048\times2048$-resolution bucket holds the minimal batch size as $B=4$. The maximum resolution is $4096\times4096$ with $B=1$. For data parallelism, each device processes distinct buckets with their corresponding batch sizes, maintaining a similar token budget.

\paragraph{Resolution-Dependent Timestep Shifting.} Sampling from a single timestep distribution is suboptimal across resolutions ranging from less than $256\times256{}$ to $4096\times4096$ pixels. Intuitively, higher-resolution images require stronger corruption (more noise) to destroy the signal, while lower-resolution images require less noise. Given an image with $n=H_1\times W_1$ pixels and its high-resolution counterpart with $m=H_2\times W_2$ pixels, \citet{esser2024sd3} provides an equation to map the the timestep $t_n$ to $t_m$:
\begin{equation}
    t_m=\frac{\sqrt{\frac{m}{n}}t_n}{1+(\sqrt{\frac{m}{n}}-1)t_n}.
    \label{eq:time_shift}
\end{equation}
In our practice, we set the base pixel number as $n=1024\times1024$, and apply this mapping to all sampled timesteps.

\paragraph{Model-Guidance Training.} \citet{tang2025dm_wo_cfg} propose a model-guidance training target to improve sampling fidelity. We adopt this approach for text-to-image training. Under our formulation, the target becomes:
\begin{equation}
\hat{\f} =  \hat{\alpha}_t\x+\hat{\sigma}_t\veps - \f_{\theta^*,t,t}^{\texttt{uncond}}) + \frac{B_{t,r}}{\frac{\dm B_{t,r}}{\dt}}\left[\frac{\dm\hat{\alpha}_t}{\dt}\cdot\x+\frac{\dm\hat{\sigma}_t}{\dt}\cdot\veps -\frac{\dm\f_{\theta^-,t,r}}{\dt}\right] + (\omega-1)(\f_{\theta^*,t,t}^{\texttt{cond}},
\label{eq:model_guidance}
\end{equation}
where $\omega$ denotes the Classifier-Free Guidance (CFG) scale, $\theta^{*}$ is the Exponential Moving Average (EMA) of $\theta$, $\f_{\theta^*,t,t}^{\texttt{cond}}$ and $\f_{\theta^*,t,t}^{\texttt{uncond}}$ respectively represent the conditional and unconditional outputs. 

\paragraph{From-Scratch Training.} The TiM-T2I model contains $865M$ parameters with the patch size of $1$. We train from scratch for about $30$ days across 16 NVIDIA-A100 GPUs with a constant learning rate of $4\times10^{-4}$, using PyTorch-FSDP~\cite{zhao2023fsdp} and half-precision ($\texttt{torch.bfloat16}$) for memory efficiency. Following \citet{tang2025dm_wo_cfg}, we use model-guidance target \cref{eq:model_guidance} with CFG scale $\omega=1.75$ after $100K$ iterations.

\section{Additional Results}~\label{appx:additional_results}

\begin{table}[ht]
\centering
\begin{adjustbox}{max width=0.93\linewidth}
\begin{tabular}{c|ccc}
\toprule
Transport & NFE=1 & NFE=8 & NFE=50 \\
\midrule
OT-FM~\cite{lipman2022flowmatching,liu2023flow} & \textbf{49.91} & 26.09 & 17.99 \\
TrigFlow~\cite{lu2024scm} & 67.32 & \textbf{25.14} & 18.28 \\
EDM~\cite{karras2022edm,karras2024edm2} & 53.64 & 37.01 & 24.06 \\
VP-SDE~\cite{song2020sde,ho2020ddpm} & 78.98 & 37.44 & 35.72 \\
\bottomrule
\end{tabular}
\end{adjustbox}
\caption{\textbf{Ablation Studies on different transports.}}
\label{tab:transports}
\vspace{-4mm}
\end{table}

\begin{table*}[t]
\centering
\begin{minipage}{0.51\textwidth}
\centering
\begin{adjustbox}{max width=1.\textwidth}
\begin{tabular}{cc|cccc}
\toprule
Method & $\eps$ & Speed & $1$-step & $4$-step & $50$-step \\
\midrule
JVP & n.a. & 1.8 $\texttt{iter}/\texttt{s}$ & 49.75 & 26.22 & 18.11 \\
\midrule
DDE & 0.0001 & 2.4 $\texttt{iter}/\texttt{s}$ & 111.25 & 23.34 & 18.38 \\
DDE & 0.0002 & 2.4 $\texttt{iter}/\texttt{s}$ & 80.14 & 23.83 & 17.58 \\
DDE & 0.0005 & 2.4 $\texttt{iter}/\texttt{s}$ & 67.09 & 24.33 & \textbf{16.93} \\
DDE & 0.001 & 2.4 $\texttt{iter}/\texttt{s}$ & \textbf{48.83} & \textbf{24.73} & 17.03 \\
DDE & 0.002 & 2.4 $\texttt{iter}/\texttt{s}$ & 49.07 & 25.54 & 17.59 \\
DDE & 0.005 & 2.4 $\texttt{iter}/\texttt{s}$ & 49.91 & 26.09 & 17.99 \\
DDE & 0.01 & 2.4 $\texttt{iter}/\texttt{s}$ & 50.05 & 26.53 & 18.33 \\
DDE & 0.02 & 2.4 $\texttt{iter}/\texttt{s}$ & 49.72 & 26.67 & 18.33 \\
DDE & 0.05 & 2.4 $\texttt{iter}/\texttt{s}$ & 49.90 & 27.05 & 18.79 \\
\bottomrule
\end{tabular}
\end{adjustbox}
\caption{\textbf{The impacts of DDE $\eps$ on generation performance.}}
\label{tab:dde_epsilon}
\end{minipage}
\begin{minipage}{0.4\textwidth}
\centering
\begin{adjustbox}{max width=1.\textwidth}
\begin{tabular}{cc|ccc}
\toprule
$t=r$ & $r=0$ & $1$-step & $4$-step & $50$-step \\
\midrule
$0\%$ & $0\%$  & 52.08 & 31.52 & 24.85 \\
$0\%$ & $10\%$  & 53.46 & 32.49 & 25.74 \\
$10\%$ & $10\%$  & 51.74 & 29.75 & 22.56 \\
$20\%$ & $10\%$  & 50.98 & 28.41 & 20.74 \\
$30\%$ & $10\%$  & 50.09 & 27.01 & 19.20 \\
$40\%$ & $10\%$  & 49.88 & 26.42 & 18.54 \\
$50\%$ & $10\%$  & \textbf{47.46} & 24.62 & 17.10 \\
$60\%$ & $10\%$  & 48.29 & 24.55 & 16.58 \\
$70\%$ & $10\%$  & 48.44 & 24.05 & 16.32 \\
$80\%$ & $10\%$  & 48.26 & \textbf{22.88} & \textbf{15.34} \\
\bottomrule
\end{tabular}
\end{adjustbox}
\caption{\textbf{Timestep sampling comparison.}}
\label{tab:dc_ratio}
\end{minipage}
\vspace{-3mm}
\end{table*}

\begin{table*}[t]
\centering
\begin{adjustbox}{max width=.98\textwidth}
\begin{tabular}{l|ccc|ccc|ccc}
\toprule
\multirow{2}*{Weighting} & 
\multicolumn{3}{c|}{$\texttt{transform}(t)=t$} & 
\multicolumn{3}{c|}{$\texttt{transform}(t)=t/(1-t)$} & 
\multicolumn{3}{c}{$\texttt{transform}(t)=\tan(t)$} 
\\
& NFE=1 & NFE=8 & NFE=50 
& NFE=1 & NFE=8 & NFE=50
& NFE=1 & NFE=8 & NFE=50
\\

\midrule
(a) Reciprocal & 48.25 & 25.29 & 17.42 & 56.65 & \textbf{24.33} & \textbf{16.58} & 49.65 & 25.22 & 17.39 \\
(b) SMS & 49.01 & 25.76 & 17.75 & 72.56 & 25.15 & 17.01 & 48.93 & 25.23 & 17.19 \\
(c) Sqrt & \textbf{48.24} & 25.82 & 17.87 & \textbf{49.93} & 24.73 & 16.85 & \textbf{47.46} & \textbf{24.62} & \textbf{17.10} \\
(d) Square & 48.55 & \textbf{25.31} & \textbf{17.11} & 70.83 & 25.79 & 17.80 & 48.86 & 24.91 & 17.15 \\

\bottomrule
\end{tabular}
\end{adjustbox}
\caption{\textbf{Ablation studies on different time weighting functions.}}
\label{tab:time_weighting}
\vspace{-3mm}
\end{table*}

\subsection{Additional Ablations}~\label{appx:additional_ablations}
We provide additional ablation results in this section.

\noindent\textbf{Transport Comparison.} We conduct ablation studies on different transports in \cref{tab:transports}. We find that different transports affect the convergence speed, where OT-FM and TrigFlow perform best, EDM is slightly worse, and VP-SDE performs the worst. Thus, we adopt OT-FM in all experiments.

\paragraph{Differential Derivation Equation Calculation.} As we incorporate a small quantity $\eps$ into \cref{eq:dde_calculation} to calculate the time derivative of network. We systematically evaluate the impact of $\eps$ on numerical accuracy in \cref{tab:dde_epsilon}and observe that $\eps\in[0.001, 0.01]$ yields high precision. For training stability, we adopt $\eps=0.005$ in all experiments.  

\paragraph{Timestep Sampling.} Using the TiM-B/4 model, we observe improved performance when a portion of timesteps is fixed to $t=r$, as in \cref{tab:dc_ratio}. The best results are obtained with $50\%$ of samples using $t=r$ (diffusion training) and $10\%$ using $r=0$ (consistency training).

\paragraph{Time Weighting.} Using the TiM-B/4 model, we provide a systematic analysis of time weighting as in \cref{eq:weight_fn}. We study three types of transformations: (1) $\texttt{transform}(t)=t$, (2) $\texttt{transform}(t)=\frac{t}{1-t}$, (3) $\texttt{transform}(t)=\tan(t)$; and four types of weighting functions: (1) Reciprocal: $\texttt{w}\_\texttt{fn}(t,r)=\frac{1}{\sigmad+t-r}$, (2) Soft-Min-SNR (SMS): $\texttt{w}\_\texttt{fn}(t,r)=\frac{1}{\sigmad^2+(t-r)^2}$, (3) SQRT: $\texttt{w}\_\texttt{fn}(t,r)=\frac{1}{\sqrt{\sigmad+t-r}}$, (4) Square: $\texttt{w}\_\texttt{fn}(t,r)=\frac{1}{(\sigmad+t-r)^2}$, where $\sigmad$ is the standard deviation of clean data $\x$ ($\sigmad=1$ in our dataset). Empirically, the combination $w(t)=$ $w(t,r)=\frac{1}{\sqrt{\sigmad+\tan(t)-\tan(r)}}$ achieves the best performance, slightly surpassing the best results reported in \cref{tab:architecture}.

\begin{table*}
\centering
\begin{adjustbox}{max width=1.\textwidth}
\begin{tabular}{l|ccc|ccccc}
\toprule
Method & Epochs & Params & NFE & \textbf{FID$\downarrow$} & \textbf{sFID$\downarrow$} & \textbf{IS$\uparrow$} & \textbf{Prec.$\uparrow$} & \textbf{Rec.$\uparrow$} \\
\midrule 
\multicolumn{9}{l}{\textbf{\textit{Generative Adversarial Networks}}}\\
BigGAN~\cite{brock2018biggan} & - & 112M & $1$ & 6.95 & 7.36 & 171.4 & 0.87 & 0.28 \\
StyleGAN-XL~\cite{sauer2022styleganxl} & - & 166M  & $1$ & 2.30  & 4.02 & 265.12 & 0.78 & 0.53 \\
GigaGAN~\cite{kang2023gigagan} & - & 569M & $1\times2$ & 3.45 & - & 225.52 & 0.84 & 0.61 \\
\midrule
\multicolumn{9}{l}{\textbf{\textit{Masked and Autoregressive Models}}}\\
Mask-GIT~\cite{chang2022maskgit} & 555 & - & - & 6.18 & - & 182.1 & - & - \\
MagViT-v2~\cite{yu2023magvitv2}   & 1080 & 307M & - & 1.78 & - & 319.4 & - & - \\
LlamaGen-XL~\cite{sun2024llamagen}  & 300  & 775M & -  & 2.62 & 5.59 & 244.08 & 0.81 & 0.58 \\
LlamaGen-XXL~\cite{sun2024llamagen}  & 300  & 1.4B & -  & 2.34 & 5.97 & 253.90 & 0.81 & 0.59 \\
LlamaGen-3B~\cite{sun2024llamagen}  & 300  & 3.1B & -  & 2.18 & 5.97 & 263.3 & 0.81 & 0.58 \\
VAR~\cite{tian2024var} & 350  & 2.0B & - & 1.80 & - & 365.4 & 0.83 & 0.57 \\
MAR~\cite{li2024mar} & 800 & 943M & - & 1.55 & -  & 303.7 & 0.81 & 0.62 \\ 
RandAR-XL~\cite{pang2025randar} & 300 & 775M & - & 2.22 & - & 314.21 & 0.80 & 0.60 \\
RandAR-XXL~\cite{pang2025randar} & 300 & 1.4B & - & 2.15 & - & 321.97 & 0.79 & 0.62 \\
\midrule
\multicolumn{9}{l}{\textbf{\textit{Multi-step Diffusion Models}}} \\
LDM-4-G~\cite{rombach2022ldm} & $170$ & 395M & $250\times2$ & 3.60 & 5.12 & 247.67 & {0.87} & 0.48 \\
SimpleDiffusion~\cite{hoogeboom2023simplediffusion}  & $800$ & 2B & $250\times2$ & 2.44 & - & 256.3 & - \\
Flag-DiT-3B$^*$~\cite{gao2024luminat2x} & $200$ & 4.23B & $250\times2$ & {1.96} & {4.43} & 284.8 & 0.82 & {0.61} \\
Large-DiT-3B$^*$~\cite{gao2024luminat2x} & $340$ & 4.23B & $250\times2$ & 2.10 & 4.52 & {304.36} & 0.82 & {0.60} \\
MDT~\cite{gao2023mdt} & $1300$ & 676M & $250\times2$ & 1.79 & 4.57 & {283.01} & 0.81 & {0.61} \\
MDTv2~\cite{gao2023mdtv2} & $700$ & 676M & $250\times2$ & 1.63 & 4.45 & 311.73 & 0.79 & {0.65} \\
DiT-XL~\cite{peebles2023dit} & $1400$ & 675M & $250\times2$ & 2.27 & 4.60 & 278.24 & 0.83 & 0.57  \\
SiT-XL~\cite{ma2024sit} & $1400$ & 675M & $250\times2$ & 2.06 & 4.49 & 277.50 & 0.83 & {0.59} \\
FlowDCN-XL~\cite{wang2024flowdcn} & $400$ & 675M & $250\times2$ & 2.00 & {4.37} & 263.16 & 0.82 & 0.58 \\
SiT-REPA-XL~\cite{yu2024repa} & $800$ & 675M & $250\times2$ & {1.42} & 4.70 & 305.7 & 0.80 & {0.65} \\
DoD-XL~\cite{yue2024dod} & $300$ & 613M & $250\times2$ & {1.73} & 5.14 & {304.31} & 0.79 & \textbf{0.64} \\
SiT-RealS-XL~\cite{xu2025reals} & $400$ & 675M & $250\times2$ & 1.82 & 4.45 & 268.54 & 0.81 & 0.60 \\
\midrule
\multicolumn{9}{l}{\textbf{\textit{Few-step Consistency Models}}} \\
\midrule
MeanFlow-XL$^\dagger$~\cite{geng2025meanflow} & $1000$ & 675M & $1$ & 3.43 & - & - & - & - \\
iCT-XL~\cite{song2023icm} & - & 675M & $1\times2$ & 20.30 & - & - & - & - \\
Shortcut-XL~\cite{frans2024shortcut} & $250$ & 675M & $1\times2$ & 10.60 & - & - & - & - \\
& & & $4\times2$ & 7.80 & - & - & - & - \\
IMM-XL~\cite{zhou2025imm} & $3840$ & 675M & $1\times2$ & 7.77 & - & - & - & - \\
& & & $2\times2$ & 3.99 & - & - & - & - \\
& & & $4\times2$ & 2.51 & - & - & - & - \\
\midrule
\multicolumn{9}{l}{\textbf{\textit{Any-step Transition Models}}} \\
\rowcolor{gray!20} 
\textbf{TiM-XL} & $300$ & 664M & $1^\dagger$ & 3.26 & 4.37 & 210.33 & 0.75 & 0.59 \\
\rowcolor{gray!20} 
& & & $1$ & 7.11 & 4.97 & 140.39 & 0.71 & 0.63 \\
\rowcolor{gray!20} 
& & & $1\times2$ & 6.14 & 6.21 & 151.79 & 0.74 & 0.59 \\
\rowcolor{gray!20} 
& & & $2\times2$ & 3.61 & 6.74 & 189.99 & 0.79 & 0.58 \\
\rowcolor{gray!20} 
& & & $4\times2$ & 2.62 & 5.57 & 203.41 & 0.79 & 0.60 \\
\rowcolor{gray!20} 
& & & $250\times2$ & 1.65 & 5.02 & 248.12 & 0.79 & 0.63 \\
\bottomrule
\end{tabular}
\end{adjustbox}
\caption{\textbf{Performance comparison on ImageNet-$256\times256$ class-guided generation.}
$^*$: Flag-DiT-3B and Large-DiT-3B actually have 4.23 billion parameters, where 3B means the parameters of all transformer blocks. $^\dagger$: means using model-guidance in the training, therefore eliminating the usage of CFG.}
\label{tab:in1k_256}
\end{table*}

\begin{table*}[t]
\centering

\begin{adjustbox}{max width=1.\textwidth}
\begin{tabular}{l|ccc|ccccc}
\toprule
Method & Epochs & Params & NFE & \textbf{FID$\downarrow$} & \textbf{sFID$\downarrow$} & \textbf{IS$\uparrow$} & \textbf{Prec.$\uparrow$} & \textbf{Rec.$\uparrow$} \\
\midrule
\multicolumn{9}{l}{\textbf{\textit{Generative Adversarial Networks}}}\\
BigGAN~\cite{brock2018biggan} & - & 160M & $1$ & 7.5 & - & 152.8 &- &-\\
StyleGAN-XL~\cite{sauer2022stylegan} & - & 168M & $1\times2$ & 2.41 & 4.06 & 267.75 & 0.77 & 0.52\\
\midrule
\multicolumn{9}{l}{\textbf{\textit{Masked and Autoregressive Models}}}\\
VAR~\cite{tian2024var} & 1080 & 307M & - & 2.63 & - & 303.2 & - & - \\
MAGVITv2~\cite{yu2023magvitv2} & 350 & 2.3B & - & 1.91 & - & 324.3 & - & - \\
\midrule
\multicolumn{9}{l}{\textbf{\textit{Multi-step Diffusion Models}}} \\
SimpleDiffusion~\cite{hoogeboom2023simplediffusion} & $800$ & 2B & $250\times2$ & 3.02 & - & 248.7 & - & - \\
DiffiT~\cite{hatamizadeh2023diffit}  & - & 561M & $250\times2$ & 2.67 & - & 252.12 & 0.83 & 0.55  \\
MaskDiT~\cite{zheng2023maskdit} & $800$ & - & $250\times2$ &{2.50} & {5.10} & 256.27 & 0.83 & {0.56} \\
Large-DiT-3B~\cite{gao2024luminat2x} & $368$ & 4.23B & $250\times2$ & 2.52 & 5.01 & {303.70} & 0.82 & 0.57 \\
ADM-G,U~\cite{dhariwal2021adm} & $400$ & 774M & $250\times2$ & 3.85 & 5.86 & 221.72 & 0.84 & 0.53 \\
U-ViT-H/2~\cite{bao2023uvit}  & $400$ & 501M & $250\times2$ & 4.05 & 6.44 & 263.79 & {0.84} & 0.48 \\
DiT-XL~\cite{peebles2023dit} & $600$ & 675M & $250\times2$ & 3.04 & {5.02} & 240.82 & \textbf{0.84} & 0.54 \\
EDM2-L~\cite{karras2024edm2} & $1468$ & 778M & $64\times2$ & 1.87 & - & - & - & - \\
EDM2-XL~\cite{karras2024edm2} & $1048$ & 1.1B & $64\times2$ & 1.80 & - & - & - & - \\
EDM2-XXL~\cite{karras2024edm2} & $734$ & 1.5B & $64\times2$ & 1.73 & - & - & - & - \\
SiT-XL~\cite{ma2024sit} & $600$ & 675M & $250\times2$ & 2.62 & {4.18} & 252.21 & {0.84} & 0.57 \\
FlowDCN-XL~\cite{wang2024flowdcn} & - & 675M & $250\times2$ & 2.44 & 4.53 & 252.8 & 0.84 & 0.54 \\
SiT-REPA-XL~\cite{yu2024repa} & $200$ & 675M & $250\times2$ & 2.08 & 4.19 & 274.6 & {0.83} & 0.58 \\
\midrule
\multicolumn{9}{l}{\textbf{\textit{Few-step Consistency Models}}} \\
sCT-L~\cite{wang2024sct} & $1273$ & 778M & $1$ & 5.15 & - & - & - & - \\
& & & $2$ & 4.65 & - & - & - & - \\
sCT-XL~\cite{wang2024sct} & $1117$ & 1.1B & $1$ & 4.33 & - & - & - & - \\
& & & $2$ & 3.73 & - & - & - & - \\
sCT-XXL~\cite{wang2024sct} & $762$ & 1.5B & $1$ & 4.29 & - & - & - & - \\
& & & $2$ & 3.76 & - & - & - & - \\
\midrule
\multicolumn{9}{l}{\textbf{\textit{Any-step Transition Models}}} \\
\rowcolor{gray!20} 
\textbf{TiM-XL} & $300$ & 664M & $1$ & 5.07 & 4.29 & 160.06 & 0.79 & 0.59 \\
\rowcolor{gray!20} 
& & & $1\times2$ & 4.79 & 4.36 & 171.73 & 0.82 & 0.57 \\
\rowcolor{gray!20} 
& & & $2\times2$ & 4.01 & 4.22 & 171.51 & 0.83 & 0.58 \\
\rowcolor{gray!20} 
& & & $4\times2$ & 2.55 & 4.24 & 207.07 & 0.83 & 0.57 \\
\rowcolor{gray!20} 
& & & $250\times2$ & 1.69 & 4.66 & 247.52 & 0.82 & 0.62 \\
\bottomrule
\end{tabular}
\end{adjustbox}
\caption{\textbf{Performance comparison on ImageNet-$512\times512$ class-guided generation.}}
\label{tab:in1k_512}
\end{table*}

\subsection{Class-Guided Image Generation}~\label{appx:additional_c2i}
We provide the results of class-guided image generation in this section.
\paragraph{Setup.} We use SD-VAE~\cite{rombach2022ldm} for ImageNet-$256\times256$ and DC-AE~\cite{chen2024dc-ae} for ImageNet-$512\times512$, with patch sizes of $2$ and $1$, respectively. We train an XL-model with $664M$ parameters for $750K$ iterations with a batch size of $512$ ($300$ epochs), using a constant learning rate of $2\times10^{-4}$ and AdamW optimizer.
We report FID~\citep{heusel2017gans}, sFID~\citep{nash2021sfid}, Inception Score (IS)~\citep{salimans2017inceptionscore}, Precision and Recall~\citep{kynk2017precisionrecall}  using ADM evaluation suite~\citep{dhariwal2021adm}. 

\paragraph{Performance Analysis.} We provide the results on ImageNet-$256\times256$ and ImageNet-$512\times512$ in \cref{tab:in1k_256,tab:in1k_512} respectively. Across both ImageNet-$256\times256$ and ImageNet-$512\times512$, TiM-XL demonstrates strong performance-efficiency trade-offs: at low NFE ($1$ to $4\times2$), it can compete with few-step consistency models, achieving comparable FID with fewer training epochs and smaller model size. When increasing NFEs, TiM-XL matches or surpasses many multi-step diffusion models in FID, despite training for only 300 epochs. Notably, TiM demonstrates remarkable generation quality and shows stable gains as NFE increases.

\section{Qualitative Results}~\label{appx:qualitative}

\begin{figure*}[t]
\begin{center}
\centerline{\includegraphics[width=\linewidth]{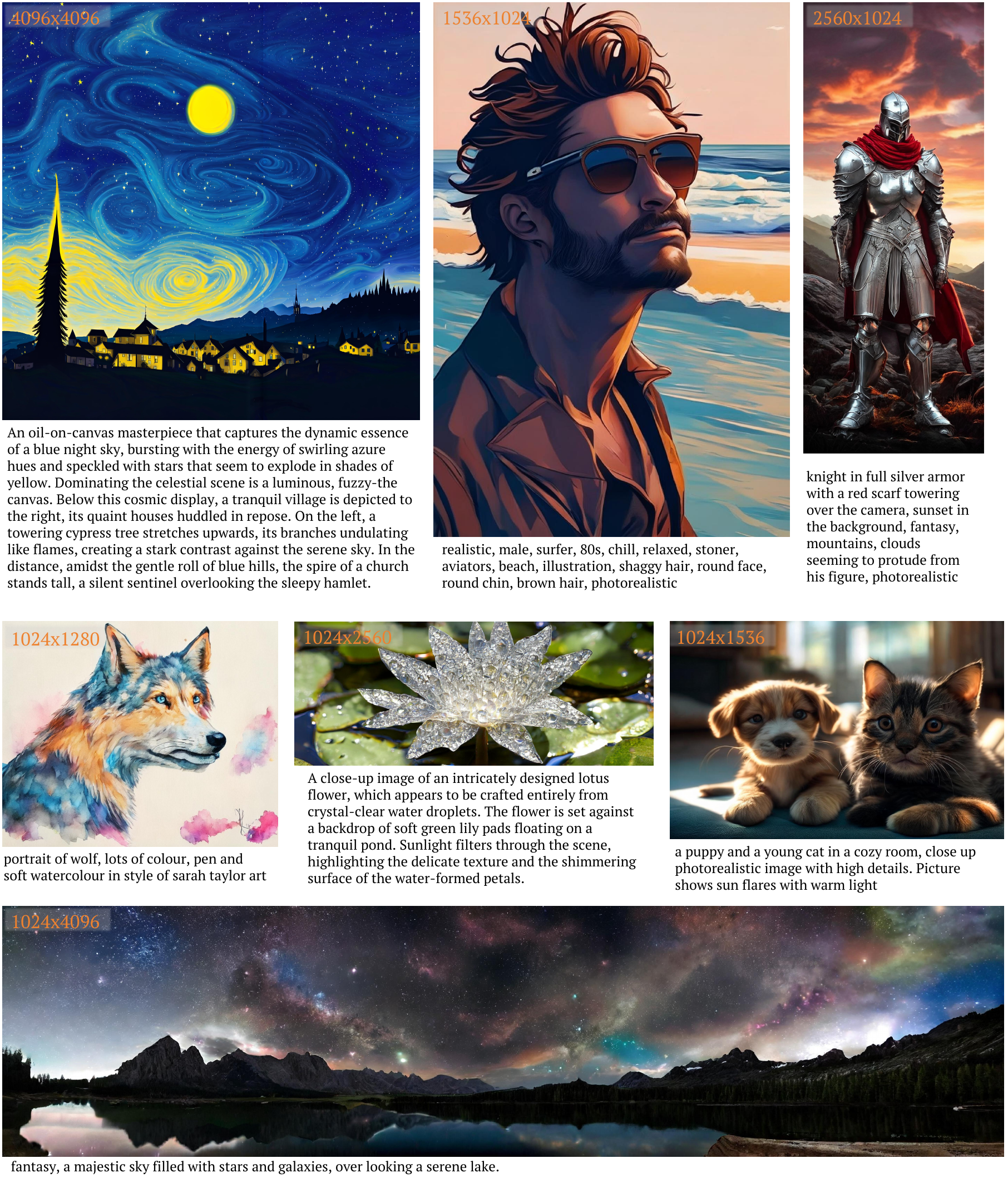}}
\caption{\textbf{High-resolution and multi-aspect generations from TiM (128 NFEs).} TiM attains up to $4096\times4096$ resolution and reliably handles multiple aspect ratios, including $1024\times4096$ and $2560\times1024$.}
\vspace{-8mm}
\label{fig:demo}
\end{center}
\end{figure*}

We provide the qualitative results in \cref{fig:demo}.

\end{document}